\documentclass[11pt]{article}
\usepackage[final]{acl}
\usepackage{times}
\usepackage{latexsym}
\usepackage{booktabs}
\usepackage{graphicx}
\usepackage{nicefrac}
\usepackage{url}
\usepackage{subcaption}
\usepackage{tabularx}
\usepackage{multirow}
\usepackage[T1]{fontenc}
\usepackage[utf8]{inputenc}
\usepackage{amsmath}
\usepackage{amsfonts}
\usepackage{pifont}
\usepackage{microtype}
\usepackage{inconsolata}

\newcommand{\cmark}{\checkmark}
\newcommand{\xmark}{\ding{55}}

\title{Script Fragmentation and Format: What Drives the English-Bengali Performance Gap in Open LLMs?}

\author{
  Shimanto Bhowmik \\ Rochester Institute of Technology \\ \texttt{sb6778@g.rit.edu}
  \And
  Tawsif Tashwar Dipto \\ Islamic University of Technology \\ \texttt{tawsiftashwar@iut-dhaka.edu}
  \AND
  Md Sazzad Islam \\ Stanford University \\ \texttt{sazzad14@stanford.edu}
  \And
  Sheryl Hsu \\ Stanford University \\ \texttt{sherylh@stanford.edu}
  \AND
  Tahsin Reasat \\ Bengali.AI \\ \texttt{reasat@bengali.ai}
}

\begin{document}

\maketitle

\begin{abstract}
Bengali is spoken by more than 230 million people, yet no standardized
instrument evaluates large language models (LLMs) on Bengali across the task
categories used to benchmark frontier models. We release 8 English benchmarks
translated into Bengali with a single consistent pipeline and use them to
evaluate 10 open LLMs from 4 families on paired English and Bengali inputs.
\emph{Script fragmentation} is what subword tokenizers do to Bengali's
alphasyllabary, whose written units are grapheme clusters spanning several
Unicode code points: they cut the script into pieces smaller than a character,
at a cost set by the vocabulary rather than the script itself. \emph{Format}
belongs to the evaluation, the answer shape that exact-match scoring demands
regardless of whether the model knew the answer. Beyond confirming a substantial
gap (macro LLM-judge score 0.79 in English versus 0.63 in Bengali), we show that
part of it is a measurement artifact: exact-match accuracy conflates correctness
with format adherence and because format failure is asymmetric across languages
it distorts the gap for some models three to five fold and even
reverses its sign for one reasoning-tuned model. On the fragmentation side,
Bengali costs roughly five times more tokens per word than English, the 10
models share only 4 vocabularies and that cost varies twofold across them: the
same Bengali word averages 7.9 tokens under the Llama~3 tokenizer but 4.0 under
Mistral's Tekken. Under three of the four vocabularies the average Bengali token
covers fewer bytes than a single code point. Bengali is
the constant here and the vocabularies are not and so what Bengali text costs is
set by tokenizer design rather than by the script. Fertility and sequence length
correlate only weakly with scores ($r = -0.23$), so we present this as a cost
and segmentation concern rather than a driver of the gap. The datasets, pipeline
and evaluation code are released.
\end{abstract}

\section{Introduction}
\label{sec:intro}

Open large language models are now routinely compared on a standard set of
English benchmarks covering commonsense reasoning, science questions,
mathematics and broad knowledge \citep{grattafiori2024llama3herdmodels,
qwen2025qwen25technicalreport}. For Bengali, one of the most widely spoken
languages in the world, no comparable evaluation suite exists. This is not
because Bengali lacks data or models: it has dedicated pretrained models such as
BanglaBERT \citep{bhattacharjee2022banglabertlanguagemodelpretraining},
BanglaGPT \citep{Salim2023BanglaGPTAG}, TituLLM
\citep{nahin2025titullmsfamilybanglallms} and TigerLLM
\citep{raihan2025tigerllmfamilybangla} and roughly 30 billion tokens of curated
pretraining text in the Sangraha corpus \citep{Khan_2024}. What Bengali lacks is
an \emph{instrument}: a set of evaluation benchmarks that spans the same task
categories as the English suites, is constructed consistently enough that scores
are comparable across tasks and comes with an analysis of where its measurements
can be trusted. Existing Bengali evaluations cover individual slices (science QA
in BEnQA \citep{SheikhShafayat2024BEnQAAQ}, the task selections of BenLLMEval
\citep{Kabir2023BenLLMEvalAC}) and translated collections such as Okapi
\citep{lai2023okapi} and Indic-Evals \citep{sarvamai2024indicevals} cover
several individual benchmarks, but no resource supports a controlled
English--Bengali comparison across the full commonsense, science, math and
multidomain spectrum. We argue that Bengali is best described not as data-poor
but as \emph{evaluation-poor} and this paper is about building and testing the
missing instrument.

The first of the two concepts is \emph{script fragmentation}, and it starts from
Bengali's writing system. Bengali is written in an alphasyllabary in which
diacritics attach on either side of a base character and consonants combine into
conjunct forms, producing multi-character grapheme clusters that constitute the
writing system's natural units \citep{alam2021large}. In the word \emph{chandra}
(moon), for example, the cluster \emph{ndra} is one written character built from
five code points: three consonants joined by two virama signs. Bengali
handwriting recognition works at this level, recovering grapheme clusters rather
than individual code points \citep{alam2021large}; subword tokenization faces
the text-side instance of the same segmentation problem. Byte-pair vocabularies
trained predominantly on Latin-script text split Bengali into units smaller than
its characters \citep{rust2021good, petrov2023tokenizer}; we measure this with
standard tokenizer statistics (\S\ref{sec:tokmetrics}). The cut runs deep: under
three of the 4 vocabularies the evaluated models share, the average Bengali
token spans 2.2 - 2.6 bytes, less than a single three-byte Bengali code point,
while the average English token spans 4.4 - 4.9 bytes. The depth is not fixed by
the script: it varies twofold across those vocabularies, so fragmentation is a
tokenizer choice. A benchmark suite with paired English and Bengali versions of
identical content is precisely the setting in which this cost can be quantified,
which is what we do here.

\emph{Format}, the second concept, is a property of the evaluation rather than
the language. Benchmark prompts instruct models to answer in a fixed shape
(\texttt{Correct Answer: [option]} in our setup) and exact-match scoring accepts
nothing else: a model that replies ``the correct answer is (b)'' or reasons
before stating the answer is scored wrong despite knowing it. The response error
rate (\S\ref{sec:metrics}) measures this failure and \S\ref{sec:format} shows it
strikes Bengali and English unequally.

Concretely, we translate 20 widely used English benchmarks into Bengali using a
single consistent GPT-4o-mini pipeline, evaluate 8 of them with 10 open
instruction-tuned models from the LLaMA, Qwen, Mistral and DeepSeek-R1 families
in the range of 3B to 72B parameters, on the paired English and Bengali versions
under three metrics: exact-match accuracy, an LLM-judge score and a
response-format error rate. The headline gap is real: macro-averaged over models
and tasks, English scores exceed Bengali scores by 0.17 - 0.18 on the 0 - 1
score scale under both metrics (LLM-judge: 0.79 vs.\ 0.63). The paired design also
lets us decompose the gap; the decomposition is where the findings lie.

Our contributions are:
\begin{itemize}
\item \textbf{A uniformly constructed Bengali evaluation suite.} 8 benchmarks
covering commonsense, science, math and multidomain knowledge, all translated
with one pipeline and released with the pipeline and evaluation code. The suite
complements existing Bengali resources: it adds WinoGrande, CommonsenseQA and
OpenBookQA, for which we are not aware of public Bengali versions and, unlike
prior translated collections assembled from heterogeneous sources, all datasets
share one pipeline, prompt format and validation procedure
(\S\ref{sec:positioning}).
\item \textbf{A paired English--Bengali evaluation of 10 open LLMs} across 4
families and three metrics, quantifying how the gap varies with model scale and
family (\S\ref{sec:results}).
\item \textbf{Evidence that exact-match scoring conflates format adherence with
capability.} When format-error rates are below 5\% in both languages,
exact-match and judge-based estimates of the cross-lingual gap agree within
about 0.05; when format errors exceed 10\%, exact match misstates the gap by
0.09 - 0.24 and can invert its sign. Reasoning-tuned models are penalized
hardest. This extends known results on format sensitivity
\citep{sclar2024quantifying, zheng2024large, alzahrani2024benchmarks} to the
cross-lingual setting, where the artifact is asymmetric and therefore biases gap
estimates rather than merely adding noise (\S\ref{sec:format}).
\item \textbf{A script-fragmentation characterization.} The script is
fixed; the tokenizer decides what its text costs. Bengali fertility averages
6.7--7.4 tokens per word against $\sim$1.3 for English, but that pooled figure
hides a twofold spread across the 4 vocabularies the 10 models share (7.9 for
Llama~3, 4.0 for Tekken); per-instance sequence lengths inflate by up to
5.5$\times$, and three of the 4 vocabularies produce tokens spanning less than a
single Bengali character while Tekken clears that boundary. We report a weak
negative association between fragmentation and scores ($r = -0.23$) and are
explicit that our evidence does not support a causal claim
(\S\ref{sec:tokenization}).
\end{itemize}

\section{Related Work}
\label{sec:related}

\paragraph{Bengali and Indic evaluation resources.}
Dedicated Bengali models include BanglaBERT
\citep{bhattacharjee2022banglabertlanguagemodelpretraining}, BanglaGPT
\citep{Salim2023BanglaGPTAG}, TituLLM \citep{nahin2025titullmsfamilybanglallms}
and TigerLLM \citep{raihan2025tigerllmfamilybangla}. On the evaluation side,
BenLLMEval assembled a first LLM evaluation for Bengali
\citep{Kabir2023BenLLMEvalAC}, BEnQA provides parallel English--Bengali science
questions from Bangladeshi curricula \citep{SheikhShafayat2024BEnQAAQ} and the
IndicLLMSuite provides pretraining and fine-tuning data across 22 Indian
languages \citep{Khan_2024}. Closest to our artifact, Okapi \citep{lai2023okapi}
released ChatGPT-translated versions of ARC, HellaSwag and MMLU in 26 languages
including Bengali, Sarvam AI's Indic-Evals collection
\citep{sarvamai2024indicevals} releases Indic translations, including Bengali,
of MMLU, BoolQ, ARC-Challenge and GSM8K and Global-MMLU
\citep{singh2025globalmmlu} includes a Bengali split with partly human-verified
translations. Section~\ref{sec:positioning} details how our suite relates to
these resources.

\paragraph{Bootstrapping benchmarks by translation.}
Translating English benchmarks is an established first step for languages
without native evaluation suites: Turkish LLM development relied on translated
benchmarks \citep{Acikgoz2024BridgingTB} and the Aya project used translation
alongside human templating across 100+ languages
\citep{ayamodelinstructionfinetuned}. The approach has known costs,
translationese, cultural mismatch and translation errors, that native
construction, as in the Khayyam Challenge \citep{ghahroodi2024khayyam}, avoids.
We take the translation route with open eyes: it is the only way to obtain
\emph{paired} English--Bengali versions of identical items, which our
measurement analysis requires (\S\ref{sec:quality}).

\paragraph{Tokenization and script fragmentation.}
Fertility and related statistics have long been used to compare tokenizers
across languages \citep{rust2021good} and tokenizer disparities translate into
real cost and context-length penalties for non-Latin scripts
\citep{petrov2023tokenizer, ahia2023languages}. Tokenizer choice measurably
affects downstream performance \citep{ali2024tokenizer,
dagan2024gettingtokenizerpretrainingdomain} and subword fragmentation degrades
robustness \citep{chai2024tokenization}. For Indic scripts specifically, recent
work proposes morphologically and orthographically grounded segmentation
\citep{brahma2025morphtok, vemula2025rethinking, libovicky2024lexically} and
byte-level modeling for Bangla \citep{bhattacharyya2025banglabyt5}. The visual
counterpart of the segmentation problem is central to Bengali handwritten
grapheme recognition \citep{alam2021large}. Our contribution is descriptive:
paired statistics locating the fragmentation below the character level.

\paragraph{Evaluation robustness and format sensitivity.}
LLM scores are highly sensitive to prompt formatting
\citep{sclar2024quantifying}, option ordering and selection biases in MCQ
\citep{zheng2024large} and small benchmark perturbations
\citep{alzahrani2024benchmarks}; prompt-sensitivity benchmarks quantify this
variance systematically \citep{razavi2025benchmarkingpromptsensitivitylarge,
Polat2024TestingPE}. Unreliable answer extraction is a recognized failure mode
of evaluation harnesses \citep{yu2025xfinder} and LLM-as-judge protocols are the
standard mitigation, with known biases of their own \citep{zheng2023judging}. We
build on this literature rather than proposing new metrics: in a cross-lingual
setting these known artifacts become \emph{asymmetric} between languages,
turning benign noise into systematic bias on gap estimates.

\section{The Bengali Benchmark Suite}
\label{sec:method}

\subsection{Benchmark Selection}
\label{sec:selection}
Starting from the suites used to report results for open model families
\citep{grattafiori2024llama3herdmodels}, we translated 20 English benchmarks
into Bengali (inventory in Appendix~\ref{app:inventory}) and selected 8 spanning
4 task categories with well-defined answers that support automatic scoring:
\textbf{commonsense and reading comprehension} (HellaSwag, WinoGrande,
CommonsenseQA, OpenBookQA, BoolQ), \textbf{science} (ARC, contributing ARC-Easy
and ARC-Challenge splits), \textbf{math} (GSM8K) and \textbf{multidomain
knowledge} (MMLU). Because ARC contributes two evaluation splits, result tables
report 9 columns for the 8 benchmarks.

\subsection{Translation Pipeline}
\label{sec:pipeline}
All datasets were translated with \texttt{gpt-4o-mini-2024-07-18}. The model was
prompted as a professional translator, instructed to preserve meaning, tone,
idioms, proper nouns and, critically, the ground-truth answer values and to emit
valid JSON so that translations could be machine-validated
(Table~\ref{tab:translation-prompt}). Automated checks validated JSON structure,
key completeness and answer-key integrity for every item; entries that failed
validation or were dropped by the batched pipeline were detected by log analysis
and retranslated until the datasets were complete. Translating all 20 benchmarks
cost approximately \$200, evidence that this bootstrap is affordable for other
underserved languages.

\subsection{Translator Selection}
\label{sec:translator}
Before committing to a translation engine, we compared Google Translate, Azure
Translator and \texttt{gpt-4o-mini-2024-07-18} in a blind review:
reviewers(native Bengali speakers) saw translations of the same source items
without knowing which system produced them and indicated which they preferred.
GPT-4o-mini was preferred for accuracy and coherence and was used for the full
pipeline. We did not log the number of reviewers, use a formal rubric or compute
inter-rater agreement, so we treat this as an informal preference test that
guided an engineering choice, not as a validation of translation quality
(\S\ref{sec:quality}).

\subsection{Position Relative to Existing Resources}
\label{sec:positioning}
Bengali versions of some of these benchmarks already exist: Okapi
\citep{lai2023okapi} includes ChatGPT-translated Bengali ARC, HellaSwag and
MMLU, the Indic-Evals collection \citep{sarvamai2024indicevals} provides Indic
translations including Bengali of MMLU, BoolQ, ARC-Challenge and GSM8K,
Global-MMLU \citep{singh2025globalmmlu} covers Bengali MMLU with partial human
verification and BEnQA \citep{SheikhShafayat2024BEnQAAQ} provides parallel
science QA. Our suite is complementary in three specific ways. First, it adds
Bengali versions of WinoGrande, CommonsenseQA and OpenBookQA, for which we are
not aware of public Bengali translations. Second, every dataset is produced by
one pipeline with one prompt, one engine and one validation procedure, so
differences between task scores reflect the tasks rather than heterogeneous
translation provenance; this consistency is what makes the cross-task and
cross-language analyses in \S\ref{sec:analysis} interpretable. Third, because
every Bengali item is paired with its English source under identical formatting,
the suite supports the controlled gap decomposition and paired tokenization
statistics that are this paper's main analytical contributions.

\subsection{Translation Quality}
\label{sec:quality}
The suite is machine-translated and has not undergone item-level human
validation; we state this up front rather than in passing. The automated
pipeline guarantees structural integrity (well-formed items, intact answer keys,
complete datasets) but not semantic fidelity. We therefore frame the suite as a
\emph{first-pass instrument with characterized error modes}: suitable for the
relative, paired comparisons performed in this paper, where translation noise
applies to all models equally and improvable through the released pipeline.
Human validation on a stratified sample is the next step; the Limitations
section returns to this point.

\section{Experimental Setup}
\label{sec:exp_details}

\subsection{Models}
We evaluate 10 open instruction-tuned models drawn from 4 families chosen to
span three axes: parameter scale (3B - 72B), documented multilingual support and
documented Bengali coverage in pretraining (Table~\ref{tab:models},
Appendix~\ref{app:repro}). The LLaMA models represent widely deployed open
models without documented Bengali pretraining, Qwen represents a multilingual
family with documented Bengali exposure, Mistral documents no Bengali support
(Mistral Small is multilingual but lists no Bengali) and the DeepSeek-R1
distillations, built on Qwen (14B) and Llama (70B) bases, represent
reasoning-tuned models, a category that matters for measurement
(\S\ref{sec:format}). All models are evaluated zero-shot, without fine-tuning,
using fixed prompt templates that instruct the model to answer in the format
\texttt{Correct Answer: [option]}; full templates and inference details are in
Appendix~\ref{app:repro}.

\subsection{Task Metrics}
\label{sec:metrics}
We score every model--dataset--language cell with three complementary metrics.

\textbf{Exact-match accuracy} counts a response as correct if it begins with the
expected answer string (after normalization such as lowercasing and label
mapping). This strictest, most common protocol is deliberately retained because
its failure modes are part of what we study.

\textbf{Response Error Rate (RER)} is the fraction of responses that do not
begin with \emph{any} valid answer option for the item:
\begin{equation}
  \mathrm{RER} = \frac{1}{n} \sum_{i=1}^{n} \mathbb{1}\!\left(\forall\,p \in P_i : \neg\,\mathrm{pre}(r_i, p)\right),
  \label{eq:rer}
\end{equation}
where $P_i$ is the set of valid (normalized) answer prefixes for item $i$ and
$\mathrm{pre}(r_i, p)$ tests whether response $r_i$ begins with $p$. Its
complement is the Response Adherence Rate,
\begin{equation}
  \mathrm{RAR} = 1 - \mathrm{RER}.
  \label{eq:rar}
\end{equation}
RER isolates \emph{format} failure from \emph{knowledge} failure: a response
with high RER cannot score under exact match regardless of whether the model
knew the answer. This makes RER the diagnostic for the answer-extraction
failures documented in evaluation-harness studies \citep{yu2025xfinder}.

\textbf{LLM-judge score} is the fraction of responses that a few-shot GPT-based
judge rules semantically equivalent to the ground truth, independent of surface
format \citep{zheng2023judging}. The judge is our estimate of capability with
format effects removed; its own reliability limits are discussed in the
Limitations section.

\subsection{Tokenization Metrics}
\label{sec:tokmetrics}
We characterize each model's tokenizer on the paired corpora using three
standard statistics, computed over every prompt (system prompt plus item) in
each dataset and language. Although we evaluate 10 models, they share only 4
distinct vocabularies: the Llama~3 tokenizer (the 4 LLaMA models and the
DeepSeek-R1 Llama distill), the Qwen~2 tokenizer (both Qwen models and the
DeepSeek-R1 Qwen distill) and the two Mistral vocabularies (Mistral 7B's 32k
SentencePiece and Mistral Small's Tekken). All tokenizer statistics therefore
characterize 4 vocabularies carried by 10 models. \textbf{Fertility} is the mean
number of tokens per whitespace-separated word \citep{rust2021good}.
\textbf{Tokens per instance} is the mean tokenized length of a full prompt, the
quantity that determines context usage and inference cost
\citep{ahia2023languages}. \textbf{Bytes per token} is the mean UTF-8 byte
length of the text divided by its token count
\citep{dagan2024gettingtokenizerpretrainingdomain}; higher values mean each
token spans more text (coarser, more efficient tokenization), lower values mean
finer fragmentation. Earlier drafts used ad hoc names for these quantities; we
use the established names for comparability with the tokenization literature.

\section{Results}
\label{sec:results}

\begin{table*}[t]
\centering
\footnotesize
\begin{tabular}{l rrr rrr rr}
\toprule
& \multicolumn{3}{c}{\textbf{Exact-match accuracy}} & \multicolumn{3}{c}{\textbf{LLM-judge}} & \multicolumn{2}{c}{\textbf{RER}} \\
\cmidrule(lr){2-4} \cmidrule(lr){5-7} \cmidrule(lr){8-9}
\textbf{Model} & EN & BN & $\Delta$ & EN & BN & $\Delta$ & EN & BN \\
\midrule
llama3.1:8b     & 0.682 & 0.401 & \underline{0.281} & 0.669 & 0.496 & \underline{0.173} & 0.020 & \textbf{0.190} \\
llama3.1:70b    & 0.885 & 0.751 & 0.133 & 0.843 & 0.748 & 0.095 & 0.007 & 0.010 \\
llama3.2:3b     & 0.658 & 0.322 & 0.336 & 0.657 & 0.353 & 0.304 & 0.017 & 0.064 \\
llama3.3:70b    & 0.851 & 0.752 & 0.099 & 0.845 & 0.745 & 0.100 & 0.048 & 0.015 \\
qwen2.5:7b      & 0.816 & 0.468 & 0.348 & 0.793 & 0.490 & 0.303 & 0.010 & 0.004 \\
qwen2.5:72b     & 0.895 & 0.599 & \underline{0.295} & 0.856 & 0.767 & \underline{0.089} & 0.009 & \textbf{0.247} \\
mistral:7b      & 0.170 & 0.112 & \underline{0.058} & 0.627 & 0.328 & \underline{0.299} & \textbf{0.733} & \textbf{0.721} \\
mistral:24b     & 0.833 & 0.637 & 0.196 & 0.806 & 0.681 & 0.125 & 0.025 & 0.066 \\
deepseek-r1:14b & 0.750 & 0.515 & \underline{0.236} & 0.908 & 0.767 & \underline{0.141} & \textbf{0.140} & \textbf{0.212} \\
deepseek-r1:70b & 0.458 & 0.638 & \underline{$-$0.179} & 0.935 & 0.879 & \underline{0.057} & \textbf{0.494} & \textbf{0.200} \\
\midrule
Mean            & 0.700 & 0.520 & 0.180 & 0.794 & 0.625 & 0.169 & 0.150 & 0.173 \\
Mean (excl.\ HellaSwag) & 0.703 & 0.540 & 0.163 & 0.830 & 0.638 & 0.192 & 0.147 & 0.157 \\
\bottomrule
\end{tabular}
\caption{Macro-averages over the 9 evaluation splits, per model and language.
$\Delta$ = EN $-$ BN. \textbf{Bold} marks RER above 0.1; for those five models
the two gap estimates (\underline{underlined}) diverge by 0.09--0.24 and can
invert in sign, while for the remaining models the two estimates agree within
0.05. The second mean row excludes HellaSwag, where the judge is unreliable
(\S\ref{sec:format}). Per-split scores are in
Tables~\ref{tab:en_bn_accuracy}--\ref{tab:en_bn_judge}.}
\label{tab:macro}
\end{table*}

\subsection{The Cross-Lingual Gap}
\begin{figure}[t]
    \centering
    \includegraphics[width=\linewidth]{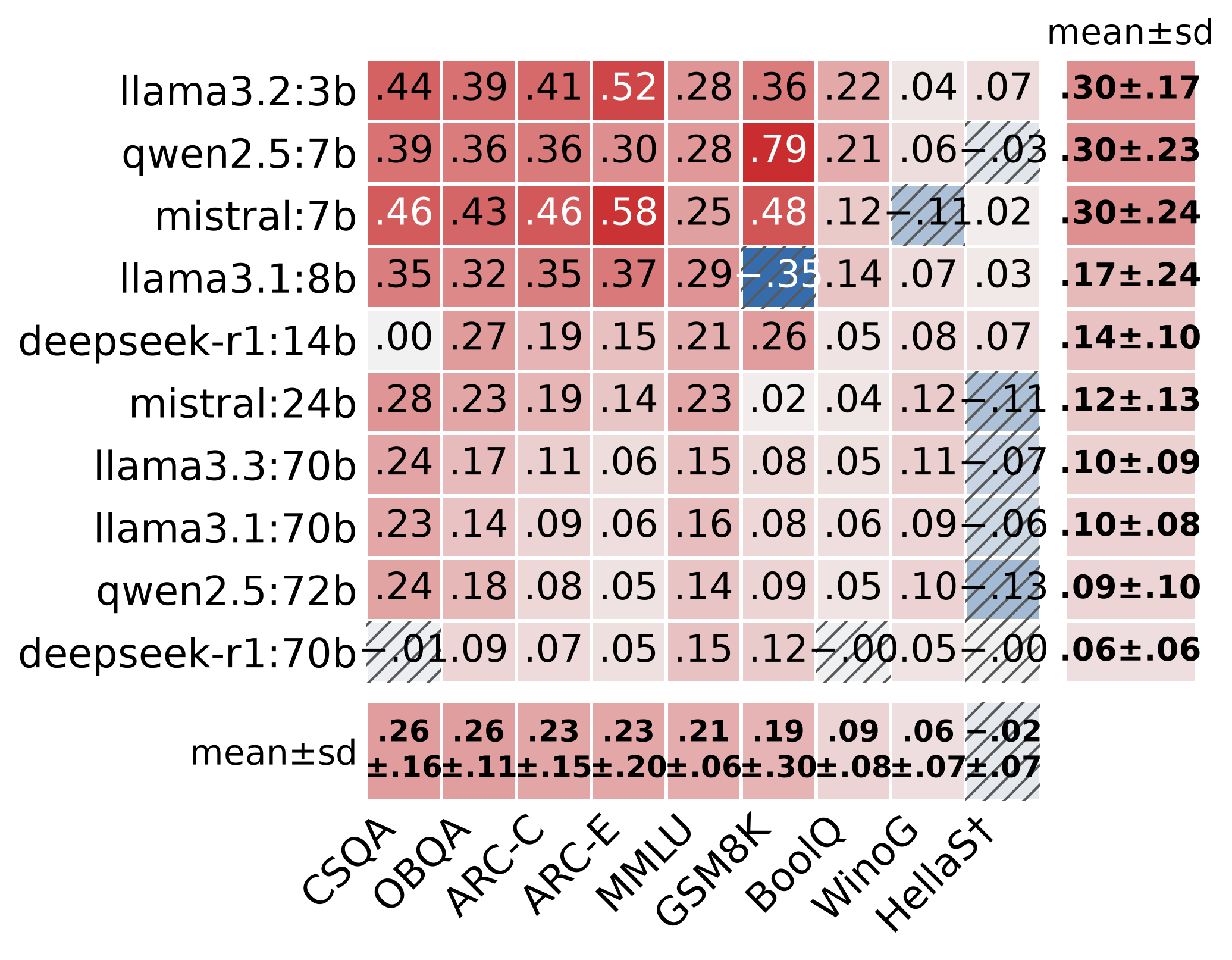}
\caption{Per-model, per-split English$-$Bengali LLM-judge gap; rows and columns
sorted by mean gap, marginals give mean$\pm$sd, hatched cells are negative
(Bengali above English) and color encodes the printed value. $\dagger$The judge
is unreliable on HellaSwag (\S\ref{sec:format}). The exact-match counterpart is
Figure~\ref{fig:gap_heatmap_em} (Appendix~\ref{app:results}).}
    \label{fig:score-diff-sorted}
\end{figure}

Table~\ref{tab:macro} summarizes performance macro-averaged over the 9
evaluation splits; Tables~\ref{tab:en_bn_accuracy}--\ref{tab:en_bn_judge} give
per-split scores. Averaged over all models, English exceeds Bengali by 0.180
under exact match and 0.169 under the LLM judge, so a substantial capability gap
survives even after format effects are removed. The gap narrows sharply with
scale within each family (row marginals of Figure~\ref{fig:score-diff-sorted}):
the LLaMA judge gap falls from 0.304 (3B) to 0.095--0.100 (70B), Qwen from 0.303
to 0.089 and Mistral from 0.299 to 0.125. The best Bengali performer overall is
deepseek-r1:70b (judge 0.879), followed by the large Qwen and LLaMA models.
Documented Bengali pretraining coverage (Table~\ref{tab:models}) shows no clear
effect at small scale: qwen2.5:7b (documented coverage) and llama3.1:8b (none)
appear to differ (judge gaps 0.303 vs.\ 0.173), but the difference vanishes once
GSM8K, a split carrying known harness anomalies (see Limitations), is excluded
entirely (0.242 vs.\ 0.239), so this pool supports no conclusion about coverage
either way.

\subsection{Where the Gap Concentrates}
Figure~\ref{fig:score-diff-sorted} maps each model's judge gap by split and two
regimes emerge: knowledge-heavy multiple choice carries the gap (CSQA, OBQA, ARC
and MMLU average 0.21--0.26) while binary BoolQ and WinoGrande stay at or below
0.09. The column marginals separate two further stories: MMLU's gap is nearly
model-independent (0.21$\pm$0.06), suggesting a property of the language pair
rather than of any model, whereas GSM8K's similar mean hides extreme model
dependence (0.19$\pm$0.30), spanning qwen2.5:7b's 0.79 collapse to llama3.1:8b's
$-$0.35. The row marginals show mean and variance falling together with scale: a
large gap is also an erratic gap. The exact-match view of the same matrix
(Figure~\ref{fig:gap_heatmap_em}, Appendix~\ref{app:results}) is instead
dominated by the format artifacts of \S\ref{sec:format}.

\section{Analysis}
\label{sec:analysis}

\subsection{Format Adherence Confounds Exact-Match Accuracy}
\label{sec:format}

\begin{figure}[t]
    \centering
    \includegraphics[width=0.78\linewidth]{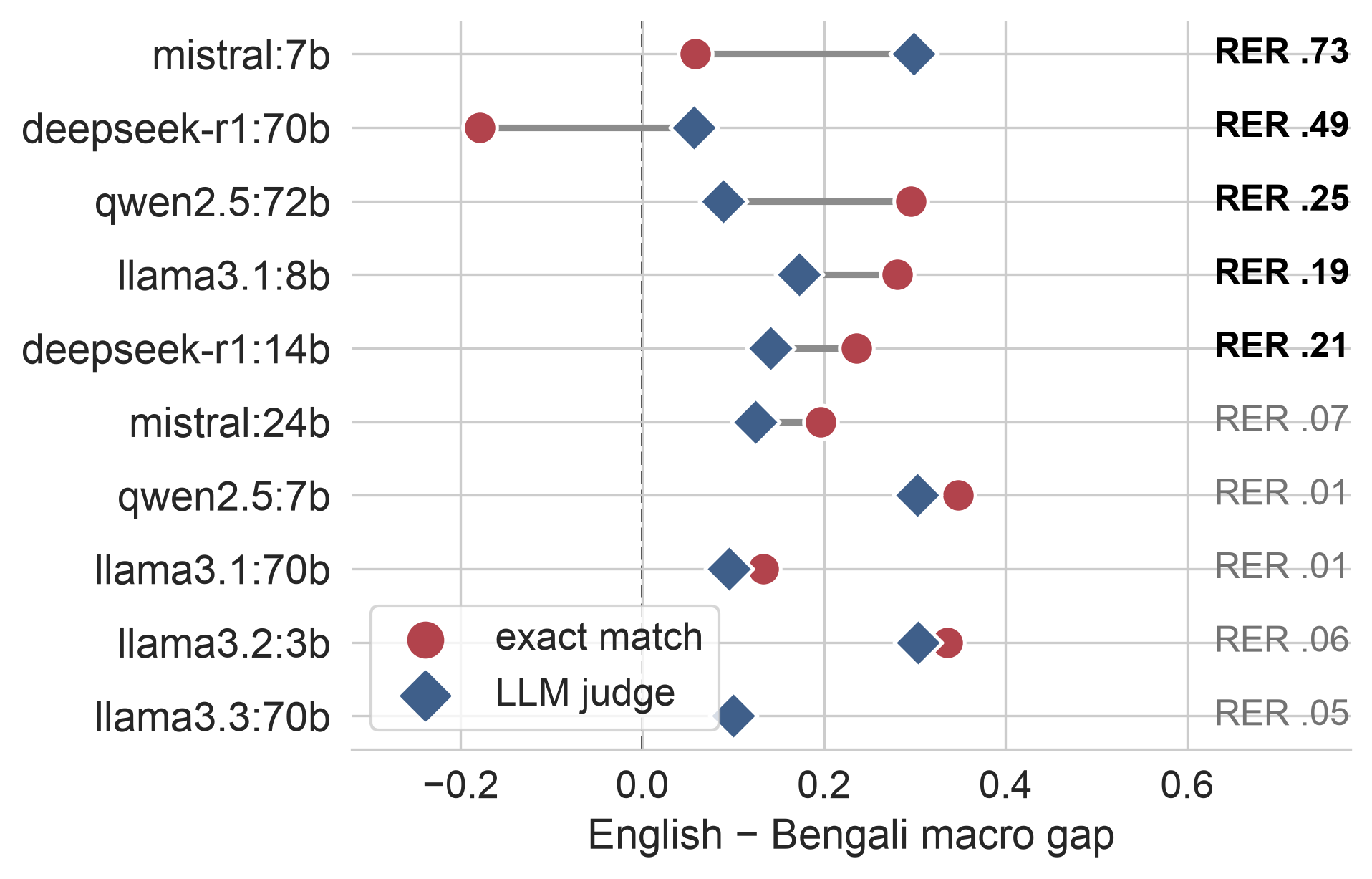}
\caption{Exact-match vs.\ LLM-judge estimates of the macro English$-$Bengali gap
per model, sorted by divergence; the right column gives each model's larger
macro RER (bold above 0.1). Format failure separates the estimates by up to 0.24
and inverts the sign for deepseek-r1:70b; near-zero-RER models (grey) agree.}
    \label{fig:format_distortion}
\end{figure}

The three metrics in Table~\ref{tab:macro} were chosen so that their
disagreements are informative and they disagree systematically. Exact-match
accuracy requires both knowing the answer and emitting it in the instructed
format; the judge requires only the former; RER measures failure of the latter
alone. Prior work established that format sensitivity adds noise to monolingual
benchmarks \citep{sclar2024quantifying, zheng2024large,
alzahrani2024benchmarks}. In our paired setting the effect is worse than noise:
RER differs \emph{between languages} for the same model, so exact match does not
merely blur the cross-lingual gap, it biases it.
\citet{SheikhShafayat2024BEnQAAQ} observe the same phenomenon qualitatively,
noting that eliciting format-conforming output is markedly harder in Bengali
than in English and fall back to manual answer assessment; our RER quantifies
that failure mode at scale. Three patterns in Tables~\ref{tab:macro}
and~\ref{tab:en_bn_rer} make the case; Figure~\ref{fig:format_distortion}
previews them.

\textbf{Apparent incapability that is actually format failure.} mistral:7b
scores 0.048 exact-match on English OpenBookQA while the judge scores it 0.678;
its RER there is 0.914, meaning nine of ten responses never began with a valid
option label. Its macro exact-match score (0.170 EN) would place it far below
every other model, while its judge score (0.627 EN) sits modestly below the
other 7--8B models (0.669--0.793) rather than at the catastrophic floor its
exact-match score implies. An earlier draft concluded that Mistral consistently
underperforms across languages; the RER column shows that conclusion was
substantially a prefix-matching artifact and we correct it here. The genuine
Mistral finding is different and appears only under the judge: proportionally,
it loses 48\% of its English judge score in Bengali (0.627 to 0.328), the
largest relative drop among the 7--8B models (38\% for qwen2.5:7b, 26\% for
llama3.1:8b), consistent with its undocumented Bengali support.

\textbf{Reasoning-tuned models are penalized hardest.} The DeepSeek-R1
distillations produce chain-of-thought before the final answer, which frequently
breaks the required answer prefix. deepseek-r1:70b has an English RER of 0.494
and an English exact-match of 0.458, against a judge score of 0.935, the highest
English judge score in the pool. Exact-match protocols thus rank the strongest
reasoner in the pool as one of the weakest models, a concrete instance of the
answer-extraction failures that motivate harness-robustness work
\citep{yu2025xfinder}.

\textbf{Language-asymmetric format failure biases the gap estimate.} Because RER
is not equal across languages, the bias does not cancel in the EN$-$BN
difference. For qwen2.5:72b, Bengali RER (0.247) far exceeds English RER (0.009)
and the exact-match gap (0.295) is more than triple the judge gap (0.089): most
of this model's apparent Bengali deficit is formatting. llama3.1:8b shows the
same pattern on OpenBookQA, where accuracy collapses from 0.790 (EN) to 0.172
(BN) while RER rises from 0.000 to 0.658. For deepseek-r1:70b the asymmetry runs
the other way (EN RER 0.494 vs.\ BN 0.200) and exact match reports a
\emph{negative} gap of $-$0.179, Bengali above English, where the judge reports
a positive gap of 0.057. Across the pool the pattern is monotone: the three
models with RER below 0.05 in both languages show gap estimates agreeing within
0.05, the two models with intermediate RER (peaking near 0.07) diverge by at
most 0.07 and the five models with RER above 0.1 in either language diverge
by 0.09 to 0.24 (deepseek-r1:14b lowest, mistral:7b highest).

\textbf{Robustness to judge errors.} The judge itself is imperfect and its
failure is systematic rather than random: on English HellaSwag, seven models
score 0.583--0.905 exact match while the judge places them between 0.352 and
0.423; five of those seven have RER at or near zero, so format failure cannot
explain the divergence and only the two chain-of-thought DeepSeek-R1 models
escape (0.898 and 0.933). Our hypothesis is response-style sensitivity in the
judge itself: HellaSwag asks for a continuation; terse label answers are judged
not equivalent to the gold continuation while reasoning models restate it in
full. The conclusions above do not depend on the split: excluding HellaSwag
(second mean row of Table~\ref{tab:macro}), qwen2.5:72b's exact-match gap
(0.247) still more than doubles its judge gap (0.116) and the deepseek-r1:70b
sign inversion persists ($-$0.188 vs.\ $+$0.064). The aggregate direction is
itself unstable: the pooled exact-match gap exceeds the judge gap with HellaSwag
included (0.180 vs.\ 0.169) and falls below it without (0.163 vs.\ 0.192); the
per-split flip is visible by comparing Figures~\ref{fig:score-diff-sorted}
and~\ref{fig:gap_heatmap_em}; the lesson is that exact match is unreliable in
both directions, with large per-model biases that partially cancel in aggregate.
The RER evidence is judge-independent throughout.

The practical conclusion for low-resource evaluation is that a single
exact-match number cannot be interpreted as a capability gap: report
format-error rates alongside accuracy and treat gap estimates as unreliable
wherever RER is non-negligible and asymmetric. This finding is invariant to our
translation-quality caveats: translation noise affects both metrics equally; the
divergences are metric-internal.

\subsection{Script Fragmentation: Fixed Terrain, Variable Cost}
\label{sec:tokenization}

The paired corpora let us measure exactly what current subword vocabularies do
to Bengali text (Figure~\ref{fig:token_stats} and additional plots in
Appendix~\ref{app:token}).

\begin{figure}[t]
    \centering
    \includegraphics[width=0.8\linewidth]{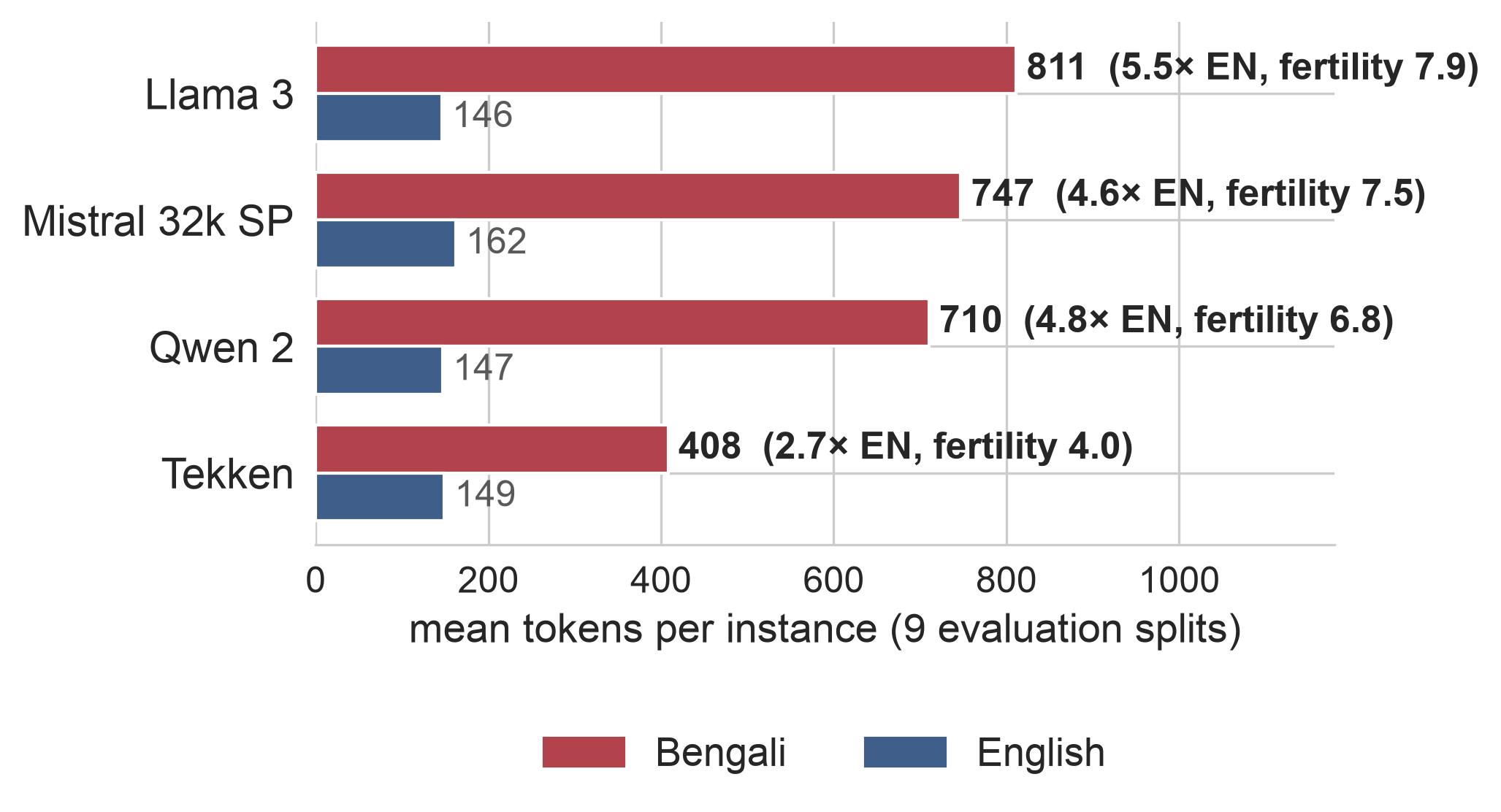}
\caption{Bengali vs.\ English tokenization cost by vocabulary: mean tokens per
instance over the 9 evaluation splits, annotated with the BN/EN ratio and
Bengali fertility. Exact values incl.\ bytes per token: Table~\ref{tab:vocab}
(Appendix~\ref{app:token}).}
    \label{fig:vocab_cost}
\end{figure}

\textbf{Per-vocabulary cost.} Figure~\ref{fig:vocab_cost} is the concrete
result: English cost is nearly identical across the 4 vocabularies (146--162
tokens per instance) while Bengali cost varies twofold. Mistral Small's Tekken
vocabulary encodes the same Bengali text in half the tokens of Llama~3 (408 vs.\
811 per instance; fertility 4.0 vs.\ 7.9), from a family with no documented
Bengali or Indic support and at a vocabulary size nearly identical to Llama~3's
(131k vs.\ 128k): vocabulary composition, not size or model scale, separates
Bengali tokenization cost across open models. The absolute cost is a direct
inference multiplier: under the dominant Llama~3 vocabulary the same content
costs 5.5 times more context in Bengali (BoolQ: 1{,}115 tokens vs.\ 196;
HellaSwag: 1{,}359 vs.\ 256), shrinking the effective context window for Bengali
users independent of any accuracy effect. Pooled across models, Bengali
fertility is 6.7--7.4 tokens per word against 1.25--1.4 for English
(Figure~\ref{fig:token_scatter_appendix}), within documented cross-lingual
disparities, which reach 15$\times$ for some languages
\citep{petrov2023tokenizer, ahia2023languages}.

\textbf{Sub-character fragmentation.} Every Bengali code point occupies 3 bytes
of UTF-8, so bytes per token tests whether a vocabulary's tokens reach the
script's characters. Three of the 4 vocabularies fall below that boundary on
Bengali text (2.2--2.6 bytes per token, Table~\ref{tab:vocab},
Appendix~\ref{app:token}): their average token covers \emph{less than one
character}. Tekken, at 4.4 bytes, is the one vocabulary that clears the
boundary, and it is the same one that halves fertility: a single mechanism,
Bengali-capable merges, explains both columns. English tokens span 4.4--4.9
bytes, four to five characters. This sub-character segmentation is the text-side
counterpart of the challenge in visual processing of Bengali script
\citep{alam2021large}; morphologically grounded Indic tokenization
\citep{brahma2025morphtok, vemula2025rethinking} and byte-level Bangla modeling
\citep{bhattacharyya2025banglabyt5} are natural remedies our suite can help
evaluate.

\textbf{Association with performance.} Across model--dataset cells, fertility
and tokens-per-instance both correlate with the LLM-judge score at $r = -0.23$
(Appendix~\ref{app:token}). We do not read this as a performance effect. The
correlation is weak, it is confounded with model family, scale and task, and it
is pseudo-replicated: the 10 models draw on only 4 vocabularies, so the
fertility values repeat and the effective sample is far smaller than 90 cells.
What the data does support are the direct measurements: the fertility premium,
the context-cost multiplier and the sub-character fragmentation, none of which
depend on any link to accuracy.

\section{Conclusion}
\label{sec:conclusion}
We released a uniformly constructed Bengali suite of 8 benchmarks, evaluated 10
open LLMs on paired English and Bengali versions and used the pairing to answer
the title's question. \emph{Format} drives the measured gap, not the real one:
language-asymmetric format failure lets exact match distort gap estimates
threefold to fivefold or invert their sign, with reasoning-tuned models
penalized most; what survives format-insensitive judging, roughly 0.17, is
real and narrows sharply with scale. \emph{Script fragmentation} is one
phenomenon with a fixed terrain and a variable cost: subword tokenizers cut
Bengali's grapheme-cluster script below the character, but how deep the cut runs
is set by the vocabulary, not the script. Bengali pays a fivefold average
fertility premium and its tokens fall below a single character under three of
the 4 vocabularies, yet the premium varies twofold by vocabulary (Tekken halves
it) and the association with scores stays weak, so vocabulary composition, not
the script itself, sets the tax. The recurring lesson: documentation is not
measurement; documented pretraining did not predict smaller gaps in this pool
and the best Bengali tokenizer came from a family documenting no Bengali
support. For practitioners: report format-error rates, corroborate exact match
with a format-insensitive metric and expect reasoning-tuned models to be
underrated by prefix matching. The suite and pipeline are a first-pass
instrument; human validation, natively authored items and grapheme-aware
tokenization come next.

\section*{Limitations}
Our datasets are machine translations that have not undergone systematic
item-level human validation, and the engine comparison in
\S\ref{sec:translator} was an informal preference test.
Translation errors, translationese and cultural mismatch may affect absolute
Bengali scores; our analyses are paired and relative, which limits this
exposure; stratified human validation is the most important future work.

We rely on a single LLM judge as our format-insensitive capability metric and do
not corroborate it with a second, independent judge. LLM judges have documented
biases and may err differently across languages \citep{zheng2023judging};
\S\ref{sec:format} documents one split (English HellaSwag) where the judge
disagrees with the gold key at zero RER, an exposure bounded by the exclusion
check and judge-independent RER evidence; human agreement over judge verdicts and
corroboration with an independent judge remain future work. A few cells score far below
documented capability (GSM8K: llama3.1:8b in English, qwen2.5:7b in Bengali),
suggesting residual harness effects; released outputs permit auditing them.

Our tokenization analysis is aggregate: we report per-vocabulary bytes per token
and fertility, but do not measure how often individual grapheme clusters are
split or relate cluster-level segmentation to per-item correctness, which we
leave to future work. Finally, the model lineup predates the newest releases:
scores age; protocols do not.

\section*{Acknowledgements}
We thank the Stanford AI Club and Phison Electronics Corporation for collaboration and computational resources.

\bibliography{citations}

@inproceedings{alam2021large,
  title={A large multi-target dataset of common bengali handwritten graphemes},
  author={Alam, Samiul and Reasat, Tahsin and Sushmit, Asif Shahriyar and Siddique, Sadi Mohammad and Rahman, Fuad and Hasan, Mahady and Humayun, Ahmed Imtiaz},
  booktitle={International Conference on Document Analysis and Recognition},
  pages={383--398},
  year={2021},
  organization={Springer}
}

@article{Kabir2023BenLLMEvalAC,
  title={BenLLM-Eval: A Comprehensive Evaluation into the Potentials and Pitfalls of Large Language Models on Bengali NLP},
  author={M. Golam Kabir and Mohammed Saidul Islam and Md Tahmid Rahman Laskar and Mir Tafseer Nayeem and M Saiful Bari and Enamul Hoque},
  journal={ArXiv},
  year={2023},
  volume={abs/2309.13173},
  url={https://api.semanticscholar.org/CorpusID:262465154}
}

@article{Salim2023BanglaGPTAG,
  title={BanglaGPT: A Generative Pretrained Transformer-Based Model for Bangla Language},
  author={Md. Shahidul Salim and Hasan Murad and Dola Das and Faisal Ahmed},
  journal={2023 International Conference on Information and Communication Technology for Sustainable Development (ICICT4SD)},
  year={2023},
  pages={56-59},
  url={https://api.semanticscholar.org/CorpusID:265056023}
}

@inproceedings{SheikhShafayat2024BEnQAAQ,
  title     = {{BEnQA}: A Question Answering Benchmark for Bengali and English},
  author    = {Shafayat, Sheikh and Hasan, H M Quamran and Mahim, Minhajur Rahman Chowdhury and Putri, Rifki Afina and Thorne, James and Oh, Alice},
  booktitle = {Findings of the Association for Computational Linguistics: ACL 2024},
  year      = {2024},
  publisher = {Association for Computational Linguistics},
  url       = {https://aclanthology.org/2024.findings-acl.68/}
}

@inproceedings{Khan_2024,
   title={IndicLLMSuite: A Blueprint for Creating Pre-training and Fine-Tuning Datasets for Indian Languages},
   url={http://dx.doi.org/10.18653/v1/2024.acl-long.843},
   DOI={10.18653/v1/2024.acl-long.843},
   booktitle={Proceedings of the 62nd Annual Meeting of the Association for Computational Linguistics (Volume 1: Long Papers)},
   publisher={Association for Computational Linguistics},
   author={Khan, Mohammed and Mehta, Priyam and Sankar, Ananth and Kumaravelan, Umashankar and Doddapaneni, Sumanth and B, Suriyaprasaad and G, Varun and Jain, Sparsh and Kunchukuttan, Anoop and Kumar, Pratyush and Dabre, Raj and Khapra, Mitesh},
   year={2024},
   pages={15831–15879} }

@inproceedings{
ghahroodi2024khayyam,
title={Khayyam Challenge (Persian{MMLU}): Is Your {LLM} Truly Wise to The Persian Language?},
author={Omid Ghahroodi and Marzia Nouri and Mohammad Vali Sanian and Alireza Sahebi and Doratossadat Dastgheib and Ehsaneddin Asgari and Mahdieh Soleymani Baghshah and Mohammad Hossein Rohban},
booktitle={First Conference on Language Modeling},
year={2024},
url={https://openreview.net/forum?id=yIEyHP7AvH}
}

@misc{ayamodelinstructionfinetuned,
      title={Aya Model: An Instruction Finetuned Open-Access Multilingual Language Model}, 
      author={Ahmet Üstün and Viraat Aryabumi and Zheng-Xin Yong and Wei-Yin Ko and Daniel D'souza and Gbemileke Onilude and Neel Bhandari and Shivalika Singh and Hui-Lee Ooi and Amr Kayid and Freddie Vargus and Phil Blunsom and Shayne Longpre and Niklas Muennighoff and Marzieh Fadaee and Julia Kreutzer and Sara Hooker},
      year={2024},
      eprint={2402.07827},
      archivePrefix={arXiv},
      primaryClass={cs.CL},
      url={https://arxiv.org/abs/2402.07827}, 
}

@article{Acikgoz2024BridgingTB,
  title={Bridging the Bosphorus: Advancing Turkish Large Language Models through Strategies for Low-Resource Language Adaptation and Benchmarking},
  author={Emre Can Acikgoz and Mete Erdogan and Deniz Yuret},
  journal={ArXiv},
  year={2024},
  volume={abs/2405.04685},
  url={https://api.semanticscholar.org/CorpusID:269626406}
}

@inproceedings{bhattacharjee2022banglabertlanguagemodelpretraining,
  title     = {{BanglaBERT}: Language Model Pretraining and Benchmarks for Low-Resource Language Understanding Evaluation in {Bangla}},
  author    = {Bhattacharjee, Abhik and Hasan, Tahmid and Ahmad, Wasi Uddin and Mubasshir, Kazi Samin and Islam, Md Saiful and Iqbal, Anindya and Rahman, M. Sohel and Shahriyar, Rifat},
  booktitle = {Findings of the Association for Computational Linguistics: NAACL 2022},
  year      = {2022},
  publisher = {Association for Computational Linguistics},
  pages     = {1318--1327},
  url       = {https://aclanthology.org/2022.findings-naacl.98/}
}

@article{Polat2024TestingPE,
  title={Testing prompt engineering methods for knowledge extraction from text},
  author={Fina Polat and Ilaria Tiddi and Paul Groth},
  journal={Semantic Web},
  year={2024},
  url={https://api.semanticscholar.org/CorpusID:272788523}
}

@misc{razavi2025benchmarkingpromptsensitivitylarge,
      title={Benchmarking Prompt Sensitivity in Large Language Models}, 
      author={Amirhossein Razavi and Mina Soltangheis and Negar Arabzadeh and Sara Salamat and Morteza Zihayat and Ebrahim Bagheri},
      year={2025},
      eprint={2502.06065},
      archivePrefix={arXiv},
      primaryClass={cs.CL},
      url={https://arxiv.org/abs/2502.06065}, 
}

@misc{grattafiori2024llama3herdmodels,
      title={The Llama 3 Herd of Models}, 
      author={Aaron Grattafiori and Abhimanyu Dubey and Abhinav Jauhri and Abhinav Pandey and Abhishek Kadian and Ahmad Al-Dahle and Aiesha Letman and Akhil Mathur and Alan Schelten and Alex Vaughan and Amy Yang and Angela Fan and Anirudh Goyal and Anthony Hartshorn and Aobo Yang and Archi Mitra and Archie Sravankumar and Artem Korenev and Arthur Hinsvark and Arun Rao and Aston Zhang and Aurelien Rodriguez and Austen Gregerson and Ava Spataru and Baptiste Roziere and Bethany Biron and Binh Tang and Bobbie Chern and Charlotte Caucheteux and Chaya Nayak and Chloe Bi and Chris Marra and Chris McConnell and Christian Keller and Christophe Touret and Chunyang Wu and Corinne Wong and Cristian Canton Ferrer and Cyrus Nikolaidis and Damien Allonsius and Daniel Song and Danielle Pintz and Danny Livshits and Danny Wyatt and David Esiobu and Dhruv Choudhary and Dhruv Mahajan and Diego Garcia-Olano and Diego Perino and Dieuwke Hupkes and Egor Lakomkin and Ehab AlBadawy and Elina Lobanova and Emily Dinan and Eric Michael Smith and Filip Radenovic and Francisco Guzmán and Frank Zhang and Gabriel Synnaeve and Gabrielle Lee and Georgia Lewis Anderson and Govind Thattai and Graeme Nail and Gregoire Mialon and Guan Pang and Guillem Cucurell and Hailey Nguyen and Hannah Korevaar and Hu Xu and Hugo Touvron and Iliyan Zarov and Imanol Arrieta Ibarra and Isabel Kloumann and Ishan Misra and Ivan Evtimov and Jack Zhang and Jade Copet and Jaewon Lee and Jan Geffert and Jana Vranes and Jason Park and Jay Mahadeokar and Jeet Shah and Jelmer van der Linde and Jennifer Billock and Jenny Hong and Jenya Lee and Jeremy Fu and Jianfeng Chi and Jianyu Huang and Jiawen Liu and Jie Wang and Jiecao Yu and Joanna Bitton and Joe Spisak and Jongsoo Park and Joseph Rocca and Joshua Johnstun and Joshua Saxe and Junteng Jia and Kalyan Vasuden Alwala and Karthik Prasad and Kartikeya Upasani and Kate Plawiak and Ke Li and Kenneth Heafield and Kevin Stone and Khalid El-Arini and Krithika Iyer and Kshitiz Malik and Kuenley Chiu and Kunal Bhalla and Kushal Lakhotia and Lauren Rantala-Yeary and Laurens van der Maaten and Lawrence Chen and Liang Tan and Liz Jenkins and Louis Martin and Lovish Madaan and Lubo Malo and Lukas Blecher and Lukas Landzaat and Luke de Oliveira and Madeline Muzzi and Mahesh Pasupuleti and Mannat Singh and Manohar Paluri and Marcin Kardas and Maria Tsimpoukelli and Mathew Oldham and Mathieu Rita and Maya Pavlova and Melanie Kambadur and Mike Lewis and Min Si and Mitesh Kumar Singh and Mona Hassan and Naman Goyal and Narjes Torabi and Nikolay Bashlykov and Nikolay Bogoychev and Niladri Chatterji and Ning Zhang and Olivier Duchenne and Onur Çelebi and Patrick Alrassy and Pengchuan Zhang and Pengwei Li and Petar Vasic and Peter Weng and Prajjwal Bhargava and Pratik Dubal and Praveen Krishnan and Punit Singh Koura and Puxin Xu and Qing He and Qingxiao Dong and Ragavan Srinivasan and Raj Ganapathy and Ramon Calderer and Ricardo Silveira Cabral and Robert Stojnic and Roberta Raileanu and Rohan Maheswari and Rohit Girdhar and Rohit Patel and Romain Sauvestre and Ronnie Polidoro and Roshan Sumbaly and Ross Taylor and Ruan Silva and Rui Hou and Rui Wang and Saghar Hosseini and Sahana Chennabasappa and Sanjay Singh and Sean Bell and Seohyun Sonia Kim and Sergey Edunov and Shaoliang Nie and Sharan Narang and Sharath Raparthy and Sheng Shen and Shengye Wan and Shruti Bhosale and Shun Zhang and Simon Vandenhende and Soumya Batra and Spencer Whitman and Sten Sootla and Stephane Collot and Suchin Gururangan and Sydney Borodinsky and Tamar Herman and Tara Fowler and Tarek Sheasha and Thomas Georgiou and Thomas Scialom and Tobias Speckbacher and Todor Mihaylov and Tong Xiao and Ujjwal Karn and Vedanuj Goswami and Vibhor Gupta and Vignesh Ramanathan and Viktor Kerkez and Vincent Gonguet and Virginie Do and Vish Vogeti and Vítor Albiero and Vladan Petrovic and Weiwei Chu and Wenhan Xiong and Wenyin Fu and Whitney Meers and Xavier Martinet and Xiaodong Wang and Xiaofang Wang and Xiaoqing Ellen Tan and Xide Xia and Xinfeng Xie and Xuchao Jia and Xuewei Wang and Yaelle Goldschlag and Yashesh Gaur and Yasmine Babaei and Yi Wen and Yiwen Song and Yuchen Zhang and Yue Li and Yuning Mao and Zacharie Delpierre Coudert and Zheng Yan and Zhengxing Chen and Zoe Papakipos and Aaditya Singh and Aayushi Srivastava and Abha Jain and Adam Kelsey and Adam Shajnfeld and Adithya Gangidi and Adolfo Victoria and Ahuva Goldstand and Ajay Menon and Ajay Sharma and Alex Boesenberg and Alexei Baevski and Allie Feinstein and Amanda Kallet and Amit Sangani and Amos Teo and Anam Yunus and Andrei Lupu and Andres Alvarado and Andrew Caples and Andrew Gu and Andrew Ho and Andrew Poulton and Andrew Ryan and Ankit Ramchandani and Annie Dong and Annie Franco and Anuj Goyal and Aparajita Saraf and Arkabandhu Chowdhury and Ashley Gabriel and Ashwin Bharambe and Assaf Eisenman and Azadeh Yazdan and Beau James and Ben Maurer and Benjamin Leonhardi and Bernie Huang and Beth Loyd and Beto De Paola and Bhargavi Paranjape and Bing Liu and Bo Wu and Boyu Ni and Braden Hancock and Bram Wasti and Brandon Spence and Brani Stojkovic and Brian Gamido and Britt Montalvo and Carl Parker and Carly Burton and Catalina Mejia and Ce Liu and Changhan Wang and Changkyu Kim and Chao Zhou and Chester Hu and Ching-Hsiang Chu and Chris Cai and Chris Tindal and Christoph Feichtenhofer and Cynthia Gao and Damon Civin and Dana Beaty and Daniel Kreymer and Daniel Li and David Adkins and David Xu and Davide Testuggine and Delia David and Devi Parikh and Diana Liskovich and Didem Foss and Dingkang Wang and Duc Le and Dustin Holland and Edward Dowling and Eissa Jamil and Elaine Montgomery and Eleonora Presani and Emily Hahn and Emily Wood and Eric-Tuan Le and Erik Brinkman and Esteban Arcaute and Evan Dunbar and Evan Smothers and Fei Sun and Felix Kreuk and Feng Tian and Filippos Kokkinos and Firat Ozgenel and Francesco Caggioni and Frank Kanayet and Frank Seide and Gabriela Medina Florez and Gabriella Schwarz and Gada Badeer and Georgia Swee and Gil Halpern and Grant Herman and Grigory Sizov and Guangyi and Zhang and Guna Lakshminarayanan and Hakan Inan and Hamid Shojanazeri and Han Zou and Hannah Wang and Hanwen Zha and Haroun Habeeb and Harrison Rudolph and Helen Suk and Henry Aspegren and Hunter Goldman and Hongyuan Zhan and Ibrahim Damlaj and Igor Molybog and Igor Tufanov and Ilias Leontiadis and Irina-Elena Veliche and Itai Gat and Jake Weissman and James Geboski and James Kohli and Janice Lam and Japhet Asher and Jean-Baptiste Gaya and Jeff Marcus and Jeff Tang and Jennifer Chan and Jenny Zhen and Jeremy Reizenstein and Jeremy Teboul and Jessica Zhong and Jian Jin and Jingyi Yang and Joe Cummings and Jon Carvill and Jon Shepard and Jonathan McPhie and Jonathan Torres and Josh Ginsburg and Junjie Wang and Kai Wu and Kam Hou U and Karan Saxena and Kartikay Khandelwal and Katayoun Zand and Kathy Matosich and Kaushik Veeraraghavan and Kelly Michelena and Keqian Li and Kiran Jagadeesh and Kun Huang and Kunal Chawla and Kyle Huang and Lailin Chen and Lakshya Garg and Lavender A and Leandro Silva and Lee Bell and Lei Zhang and Liangpeng Guo and Licheng Yu and Liron Moshkovich and Luca Wehrstedt and Madian Khabsa and Manav Avalani and Manish Bhatt and Martynas Mankus and Matan Hasson and Matthew Lennie and Matthias Reso and Maxim Groshev and Maxim Naumov and Maya Lathi and Meghan Keneally and Miao Liu and Michael L. Seltzer and Michal Valko and Michelle Restrepo and Mihir Patel and Mik Vyatskov and Mikayel Samvelyan and Mike Clark and Mike Macey and Mike Wang and Miquel Jubert Hermoso and Mo Metanat and Mohammad Rastegari and Munish Bansal and Nandhini Santhanam and Natascha Parks and Natasha White and Navyata Bawa and Nayan Singhal and Nick Egebo and Nicolas Usunier and Nikhil Mehta and Nikolay Pavlovich Laptev and Ning Dong and Norman Cheng and Oleg Chernoguz and Olivia Hart and Omkar Salpekar and Ozlem Kalinli and Parkin Kent and Parth Parekh and Paul Saab and Pavan Balaji and Pedro Rittner and Philip Bontrager and Pierre Roux and Piotr Dollar and Polina Zvyagina and Prashant Ratanchandani and Pritish Yuvraj and Qian Liang and Rachad Alao and Rachel Rodriguez and Rafi Ayub and Raghotham Murthy and Raghu Nayani and Rahul Mitra and Rangaprabhu Parthasarathy and Raymond Li and Rebekkah Hogan and Robin Battey and Rocky Wang and Russ Howes and Ruty Rinott and Sachin Mehta and Sachin Siby and Sai Jayesh Bondu and Samyak Datta and Sara Chugh and Sara Hunt and Sargun Dhillon and Sasha Sidorov and Satadru Pan and Saurabh Mahajan and Saurabh Verma and Seiji Yamamoto and Sharadh Ramaswamy and Shaun Lindsay and Shaun Lindsay and Sheng Feng and Shenghao Lin and Shengxin Cindy Zha and Shishir Patil and Shiva Shankar and Shuqiang Zhang and Shuqiang Zhang and Sinong Wang and Sneha Agarwal and Soji Sajuyigbe and Soumith Chintala and Stephanie Max and Stephen Chen and Steve Kehoe and Steve Satterfield and Sudarshan Govindaprasad and Sumit Gupta and Summer Deng and Sungmin Cho and Sunny Virk and Suraj Subramanian and Sy Choudhury and Sydney Goldman and Tal Remez and Tamar Glaser and Tamara Best and Thilo Koehler and Thomas Robinson and Tianhe Li and Tianjun Zhang and Tim Matthews and Timothy Chou and Tzook Shaked and Varun Vontimitta and Victoria Ajayi and Victoria Montanez and Vijai Mohan and Vinay Satish Kumar and Vishal Mangla and Vlad Ionescu and Vlad Poenaru and Vlad Tiberiu Mihailescu and Vladimir Ivanov and Wei Li and Wenchen Wang and Wenwen Jiang and Wes Bouaziz and Will Constable and Xiaocheng Tang and Xiaojian Wu and Xiaolan Wang and Xilun Wu and Xinbo Gao and Yaniv Kleinman and Yanjun Chen and Ye Hu and Ye Jia and Ye Qi and Yenda Li and Yilin Zhang and Ying Zhang and Yossi Adi and Youngjin Nam and Yu and Wang and Yu Zhao and Yuchen Hao and Yundi Qian and Yunlu Li and Yuzi He and Zach Rait and Zachary DeVito and Zef Rosnbrick and Zhaoduo Wen and Zhenyu Yang and Zhiwei Zhao and Zhiyu Ma},
      year={2024},
      eprint={2407.21783},
      archivePrefix={arXiv},
      primaryClass={cs.AI},
      url={https://arxiv.org/abs/2407.21783}, 
}

@misc{qwen2025qwen25technicalreport,
      title={Qwen2.5 Technical Report}, 
      author={An Yang and Baosong Yang and Beichen Zhang and Binyuan Hui and Bo Zheng and Bowen Yu and Chengyuan Li and Dayiheng Liu and Fei Huang and Haoran Wei and Huan Lin and Jian Yang and Jianhong Tu and Jianwei Zhang and Jianxin Yang and Jiaxi Yang and Jingren Zhou and Junyang Lin and Kai Dang and Keming Lu and Keqin Bao and Kexin Yang and Le Yu and Mei Li and Mingfeng Xue and Pei Zhang and Qin Zhu and Rui Men and Runji Lin and Tianhao Li and Tianyi Tang and Tingyu Xia and Xingzhang Ren and Xuancheng Ren and Yang Fan and Yang Su and Yichang Zhang and Yu Wan and Yuqiong Liu and Zeyu Cui and Zhenru Zhang and Zihan Qiu},
      year={2025},
      eprint={2412.15115},
      archivePrefix={arXiv},
      primaryClass={cs.CL},
      url={https://arxiv.org/abs/2412.15115}, 
}

@misc{dagan2024gettingtokenizerpretrainingdomain,
      title={Getting the most out of your tokenizer for pre-training and domain adaptation}, 
      author={Gautier Dagan and Gabriel Synnaeve and Baptiste Rozière},
      year={2024},
      eprint={2402.01035},
      archivePrefix={arXiv},
      primaryClass={cs.CL},
      url={https://arxiv.org/abs/2402.01035}, 
}

@article{guo2025deepseek,
  title={Deepseek-r1: Incentivizing reasoning capability in llms via reinforcement learning},
  author={Guo, Daya and Yang, Dejian and Zhang, Haowei and Song, Junxiao and Zhang, Ruoyu and Xu, Runxin and Zhu, Qihao and Ma, Shirong and Wang, Peiyi and Bi, Xiao and others},
  journal={arXiv preprint arXiv:2501.12948},
  year={2025}
}

@misc{jiang2023mistral7b,
      title={Mistral 7B}, 
      author={Albert Q. Jiang and Alexandre Sablayrolles and Arthur Mensch and Chris Bamford and Devendra Singh Chaplot and Diego de las Casas and Florian Bressand and Gianna Lengyel and Guillaume Lample and Lucile Saulnier and Lélio Renard Lavaud and Marie-Anne Lachaux and Pierre Stock and Teven Le Scao and Thibaut Lavril and Thomas Wang and Timothée Lacroix and William El Sayed},
      year={2023},
      eprint={2310.06825},
      archivePrefix={arXiv},
      primaryClass={cs.CL},
      url={https://arxiv.org/abs/2310.06825}, 
}

@misc{mistral-small-24b-instruct-2501,
  title        = {{Mistral-Small-24B-Instruct-2501}},
  author       = {{Mistral AI Team}},
  year         = {2025},
  howpublished = {Hugging Face Model Card, Apache-2.0 licensed},
  note         = {\url{https://huggingface.co/mistralai/Mistral-Small-24B-Instruct-2501} (accessed 2025-07-20)}
}

@misc{nahin2025titullmsfamilybanglallms,
      title={TituLLMs: A Family of Bangla LLMs with Comprehensive Benchmarking}, 
      author={Shahriar Kabir Nahin and Rabindra Nath Nandi and Sagor Sarker and Quazi Sarwar Muhtaseem and Md Kowsher and Apu Chandraw Shill and Md Ibrahim and Mehadi Hasan Menon and Tareq Al Muntasir and Firoj Alam},
      year={2025},
      eprint={2502.11187},
      archivePrefix={arXiv},
      primaryClass={cs.CL},
      url={https://arxiv.org/abs/2502.11187}, 
}

@misc{raihan2025tigerllmfamilybangla,
      title={TigerLLM - A Family of Bangla Large Language Models}, 
      author={Nishat Raihan and Marcos Zampieri},
      year={2025},
      eprint={2503.10995},
      archivePrefix={arXiv},
      primaryClass={cs.CL},
      url={https://arxiv.org/abs/2503.10995}, 
}

@inproceedings{chai2024tokenization,
  title     = {Tokenization Falling Short: On Subword Robustness in Large Language Models},
  author    = {Chai, Yekun and Fang, Yewei and Peng, Qiwei and Li, Xuhong},
  booktitle = {Findings of the Association for Computational Linguistics: EMNLP 2024},
  year      = {2024},
  address   = {Miami, Florida, USA},
  publisher = {Association for Computational Linguistics},
  pages     = {1582--1599},
  url       = {https://aclanthology.org/2024.findings-emnlp.86/}
}

@inproceedings{ali2024tokenizer,
  title     = {Tokenizer Choice For {LLM} Training: Negligible or Crucial?},
  author    = {Ali, Mehdi and Fromm, Michael and Thellmann, Klaudia and Rutmann, Richard and L{\"u}bbering, Max and Leveling, Johannes and others},
  booktitle = {Findings of the Association for Computational Linguistics: NAACL 2024},
  year      = {2024},
  address   = {Mexico City, Mexico},
  publisher = {Association for Computational Linguistics},
  pages     = {3907--3924},
  url       = {https://aclanthology.org/2024.findings-naacl.247/}
}

@inproceedings{libovicky2024lexically,
  title     = {Lexically Grounded Subword Segmentation},
  author    = {Libovick{\'y}, Jind{\v{r}}ich and Helcl, Jind{\v{r}}ich},
  booktitle = {Proceedings of the 2024 Conference on Empirical Methods in Natural Language Processing},
  year      = {2024},
  address   = {Miami, Florida, USA},
  publisher = {Association for Computational Linguistics},
  pages     = {7403--7420},
  url       = {https://aclanthology.org/2024.emnlp-main.421/}
}

@misc{brahma2025morphtok,
  title  = {{MorphTok}: Morphologically Grounded Tokenization for {Indian} Languages},
  author = {Brahma, Maharaj and Karthika, N J and Singh, Atul and Adiga, Devaraj and Bhate, Smruti and Ramakrishnan, Ganesh and Saluja, Rohit and Desarkar, Maunendra Sankar},
  year   = {2025},
  eprint = {2504.10335},
  archivePrefix = {arXiv},
  primaryClass  = {cs.CL},
  url    = {https://arxiv.org/abs/2504.10335}
}

@misc{vemula2025rethinking,
  title  = {Rethinking Tokenization for Rich Morphology: The Dominance of Unigram over {BPE} and Morphological Alignment},
  author = {Vemula, Saketh Reddy and Dandapat, Sandipan and Sharma, Dipti Misra and Krishnamurthy, Parameswari},
  year   = {2025},
  eprint = {2508.08424},
  archivePrefix = {arXiv},
  primaryClass  = {cs.CL},
  url    = {https://arxiv.org/abs/2508.08424}
}

@misc{bhattacharyya2025banglabyt5,
  title  = {{BanglaByT5}: Byte-Level Modelling for {Bangla}},
  author = {Bhattacharyya, Pramit and Bhattacharya, Arnab},
  year   = {2025},
  eprint = {2505.17102},
  archivePrefix = {arXiv},
  primaryClass  = {cs.CL},
  url    = {https://arxiv.org/abs/2505.17102}
}

@misc{sarvamai2024indicevals,
  title        = {Indic-Evals: Translated Versions of Popular {LLM} Benchmarks},
  author       = {{Sarvam AI}},
  year         = {2024},
  howpublished = {Hugging Face collection},
  url          = {https://huggingface.co/collections/sarvamai/indic-evals-67196d8d0edc751606d8b2e9},
  note         = {Includes MMLU, BoolQ, ARC-Challenge, GSM8K, and TriviaQA-MCQ translated into 10 Indian languages including Bengali}
}

@inproceedings{petrov2023tokenizer,
  title     = {Language Model Tokenizers Introduce Unfairness Between Languages},
  author    = {Petrov, Aleksandar and La Malfa, Emanuele and Torr, Philip and Bibi, Adel},
  booktitle = {Advances in Neural Information Processing Systems 36 (NeurIPS 2023)},
  year      = {2023},
  url       = {https://proceedings.neurips.cc/paper_files/paper/2023/hash/74bb24dca8334adce292883b4b651eda-Abstract-Conference.html}
}

@inproceedings{rust2021good,
  title     = {How Good is Your Tokenizer? On the Monolingual Performance of Multilingual Language Models},
  author    = {Rust, Phillip and Pfeiffer, Jonas and Vuli{\'c}, Ivan and Ruder, Sebastian and Gurevych, Iryna},
  booktitle = {Proceedings of the 59th Annual Meeting of the Association for Computational Linguistics and the 11th International Joint Conference on Natural Language Processing (Volume 1: Long Papers)},
  year      = {2021},
  address   = {Online},
  publisher = {Association for Computational Linguistics},
  pages     = {3118--3135},
  url       = {https://aclanthology.org/2021.acl-long.243/}
}

@inproceedings{sclar2024quantifying,
  title     = {Quantifying Language Models' Sensitivity to Spurious Features in Prompt Design or: How {I} Learned to Start Worrying about Prompt Formatting},
  author    = {Sclar, Melanie and Choi, Yejin and Tsvetkov, Yulia and Suhr, Alane},
  booktitle = {The Twelfth International Conference on Learning Representations (ICLR)},
  year      = {2024},
  url       = {https://arxiv.org/abs/2310.11324}
}

@inproceedings{zheng2024large,
  title     = {Large Language Models Are Not Robust Multiple Choice Selectors},
  author    = {Zheng, Chujie and Zhou, Hao and Meng, Fandong and Zhou, Jie and Huang, Minlie},
  booktitle = {The Twelfth International Conference on Learning Representations (ICLR)},
  year      = {2024},
  url       = {https://arxiv.org/abs/2309.03882}
}

@inproceedings{alzahrani2024benchmarks,
  title     = {When Benchmarks are Targets: Revealing the Sensitivity of Large Language Model Leaderboards},
  author    = {Alzahrani, Norah and Alyahya, Hisham Abdullah and Alnumay, Yazeed and Alrashed, Sultan and Alsubaie, Shaykhah and Almushaykeh, Yusef and others},
  booktitle = {Proceedings of the 62nd Annual Meeting of the Association for Computational Linguistics (Volume 1: Long Papers)},
  year      = {2024},
  address   = {Bangkok, Thailand},
  publisher = {Association for Computational Linguistics},
  pages     = {13787--13805},
  url       = {https://aclanthology.org/2024.acl-long.744/}
}

@inproceedings{zheng2023judging,
  title     = {Judging {LLM}-as-a-Judge with {MT-Bench} and {Chatbot Arena}},
  author    = {Zheng, Lianmin and Chiang, Wei-Lin and Sheng, Ying and Zhuang, Siyuan and Wu, Zhanghao and Zhuang, Yonghao and others},
  booktitle = {Advances in Neural Information Processing Systems 36 (NeurIPS 2023), Datasets and Benchmarks Track},
  year      = {2023},
  url       = {https://arxiv.org/abs/2306.05685}
}

@inproceedings{ahia2023languages,
  title     = {Do All Languages Cost the Same? Tokenization in the Era of Commercial Language Models},
  author    = {Ahia, Orevaoghene and Kumar, Sachin and Gonen, Hila and Kasai, Jungo and Mortensen, David and Smith, Noah and Tsvetkov, Yulia},
  booktitle = {Proceedings of the 2023 Conference on Empirical Methods in Natural Language Processing},
  year      = {2023},
  address   = {Singapore},
  publisher = {Association for Computational Linguistics},
  pages     = {9904--9923},
  url       = {https://aclanthology.org/2023.emnlp-main.614/}
}

@inproceedings{yu2025xfinder,
  title     = {{xFinder}: Large Language Models as Automated Evaluators for Reliable Evaluation},
  author    = {Yu, Qingchen and Zheng, Zifan and Song, Shichao and Li, Zhiyu and Xiong, Feiyu and Tang, Bo and Chen, Ding},
  booktitle = {The Thirteenth International Conference on Learning Representations (ICLR)},
  year      = {2025},
  url       = {https://arxiv.org/abs/2405.11874}
}

@inproceedings{lai2023okapi,
  title     = {Okapi: Instruction-tuned Large Language Models in Multiple Languages with Reinforcement Learning from Human Feedback},
  author    = {Lai, Viet and Nguyen, Chien and Ngo, Nghia and Nguyen, Thuat and Dernoncourt, Franck and Rossi, Ryan and Nguyen, Thien},
  booktitle = {Proceedings of the 2023 Conference on Empirical Methods in Natural Language Processing: System Demonstrations},
  year      = {2023},
  address   = {Singapore},
  publisher = {Association for Computational Linguistics},
  pages     = {318--327},
  url       = {https://aclanthology.org/2023.emnlp-demo.28/}
}

@inproceedings{singh2025globalmmlu,
  title     = {Global {MMLU}: Understanding and Addressing Cultural and Linguistic Biases in Multilingual Evaluation},
  author    = {Singh, Shivalika and Romanou, Angelika and Fourrier, Cl{\'e}mentine and Adelani, David Ifeoluwa and Ngui, Jian Gang and Vila-Suero, Daniel and others},
  booktitle = {Proceedings of the 63rd Annual Meeting of the Association for Computational Linguistics (Volume 1: Long Papers)},
  year      = {2025},
  publisher = {Association for Computational Linguistics},
  url       = {https://aclanthology.org/2025.acl-long.919/}
}

\appendix

\section{Translated Benchmark Inventory}
\label{app:inventory}
Table~\ref{tab:inventory} lists all 20 benchmarks translated by the pipeline.
The 8 evaluated in this paper are OpenBookQA, ARC (Easy/Challenge), BoolQ,
CommonsenseQA, GSM8K, HellaSwag, WinoGrande and MMLU; the remainder are released
for future work.

\begin{table*}[t]
\centering
\small
\begin{tabular}{l rrr l l}
\toprule
Dataset & Train & Dev & Test & Answer format & Focus \\
\midrule
OpenBookQA       & 4957   & 500   & 500    & MCQ              & Multi-step commonsense reasoning \\
ARC              & 3370   & 869   & 3548   & MCQ              & Grade-school science \\
BoolQ            & 9427   & 3270  & --     & Yes/No           & Reading comprehension \\
CommonsenseQA    & 9741   & 1221  & 1140   & MCQ              & Commonsense reasoning \\
GSM8K            & 7473   & --    & 1319   & Number           & Grade-school math \\
HellaSwag        & 39905  & 10042 & 10003  & MCQ              & Commonsense continuation \\
WinoGrande       & 19482  & 1267  & 1767   & MCQ              & Pronoun resolution \\
MMLU             & 98487  & 1528  & 13869  & MCQ              & Multidomain knowledge \\
\midrule
BigBenchHard     & --     & --    & var.   & MCQ              & Logical reasoning \\
AlpacaEval       & --     & --    & 10465  & Instruction      & Instruction following \\
Anthropic HH     & 86372  & --    & 35006  & Free-form        & Safety, helpfulness \\
APPS             & 5000   & --    & 5000   & Code             & Coding \\
BFCL             & --     & --    & 250    & Structured       & Function calling \\
Dolly            & --     & --    & 7295   & Instruction      & Varied NLP tasks \\
HumanEval        & --     & --    & 164    & Code             & Code generation \\
MATH             & 8599   & --    & 4999   & Exact match      & Competition math \\
MMLU-Pro         & --     & 70    & 12032  & MCQ              & Multidomain knowledge \\
MR-GSM8K         & --     & --    & 12024  & Exact match      & Math meta-reasoning \\
PIQA             & 16113  & --    & 3084   & MCQ              & Physical commonsense \\
SIQA             & 33410  & 1954  & --     & MCQ              & Social reasoning \\
TruthfulQA       & --     & --    & 1634   & MCQ              & Truthfulness \\
\bottomrule
\end{tabular}
\caption{All translated benchmarks. The top block lists the 8 evaluated in this
paper; the bottom block lists the 12 additional translated benchmarks released
for future work.}
\label{tab:inventory}
\end{table*}

\section{Additional Tokenization Results}
\label{app:token}

\begin{table}[t]
\centering
\scriptsize
\setlength{\tabcolsep}{3pt}
\begin{tabular}{l l rrrrr}
\toprule
Vocabulary & Models & BN & EN & BN/EN & Fert. & B/tok \\
\midrule
Llama 3        & 4 LLaMA, R1-70B & 811 & 146 & 5.5 & 7.9 & 2.2 \\
Qwen 2         & 2 Qwen, R1-14B  & 710 & 147 & 4.8 & 6.8 & 2.6 \\
Mistral 32k SP & Mistral 7B      & 747 & 162 & 4.6 & 7.5 & 2.3 \\
Tekken         & Mistral Small   & 408 & 149 & 2.7 & 4.0 & 4.4 \\
\bottomrule
\end{tabular}
\caption{Per-vocabulary tokenization of Bengali. Tokens per instance (BN, EN,
ratio) are pipeline-computed over the full evaluation corpora; Bengali fertility
(tokens per word) and bytes per token are measured by direct tokenization of
Bengali text with each public tokenizer and agree with the pipeline statistics
(Appendix~\ref{app:token}). A Bengali code point is 3 bytes: only Tekken's
tokens clear the character boundary.}
\label{tab:vocab}
\end{table}

\begin{figure*}[t]
    \centering
    \begin{subfigure}[t]{0.40\textwidth}
        \centering
        \includegraphics[width=\linewidth]{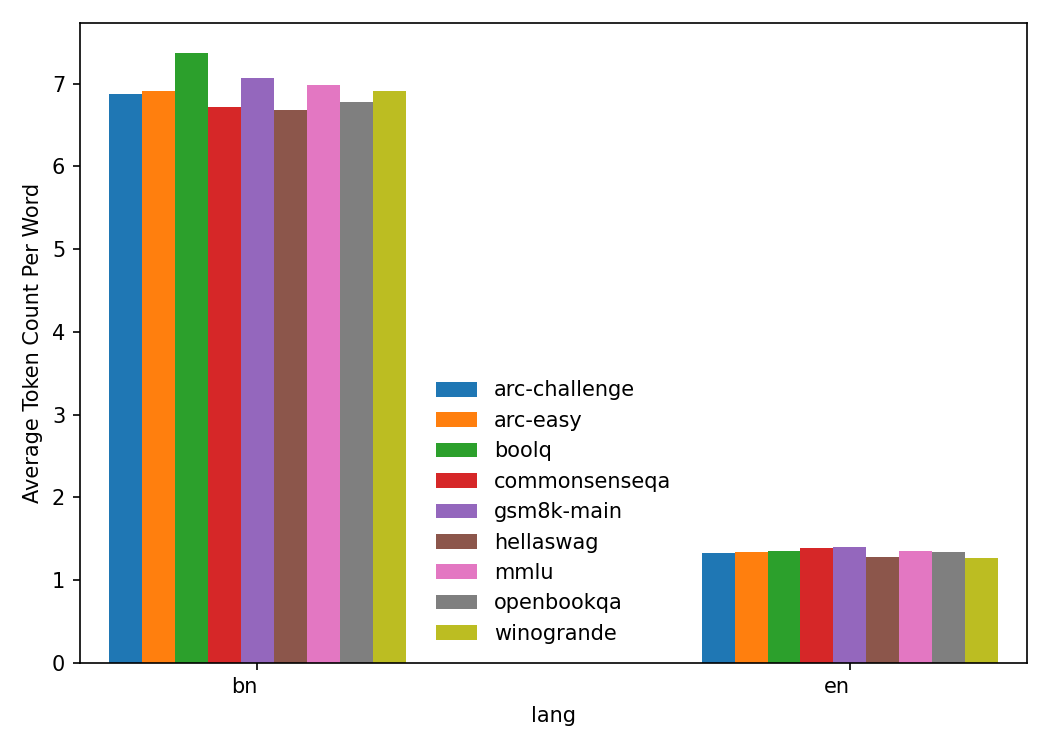}
\caption{Fertility (tokens per word).}
        \label{fig:fertility}
    \end{subfigure}
    \hfill
    \begin{subfigure}[t]{0.40\textwidth}
        \centering
        \includegraphics[width=\linewidth]{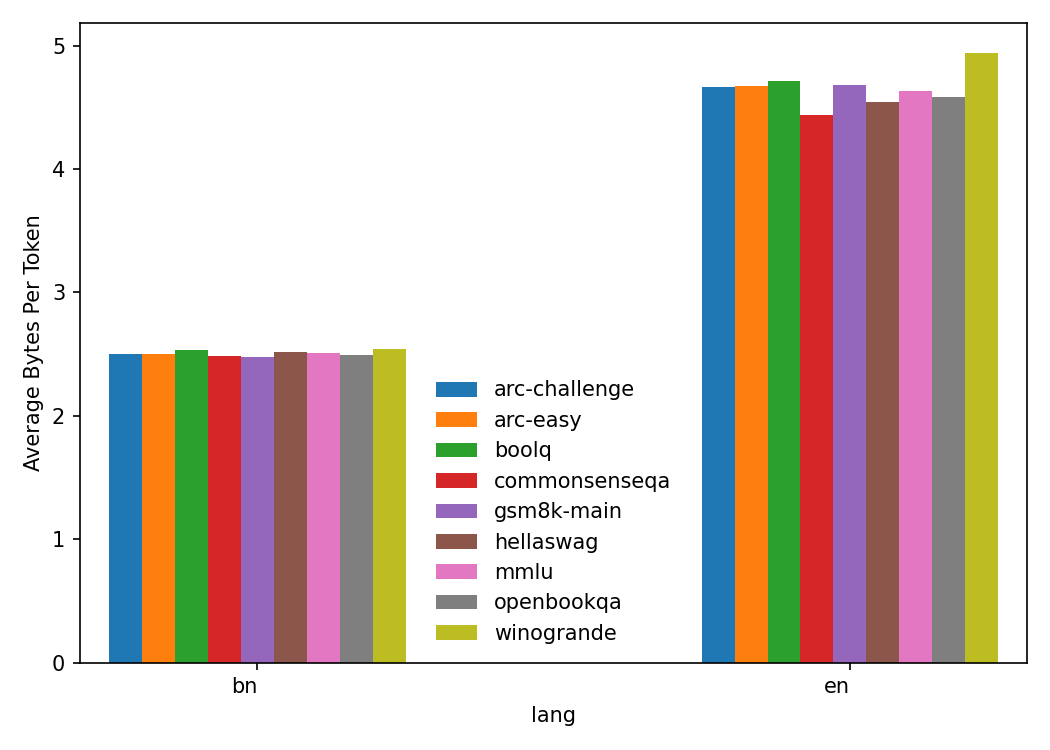}
\caption{Bytes per token.}
        \label{fig:bytes_per_token}
    \end{subfigure}
\caption{Paired tokenizer statistics by dataset and language. Bengali fertility
is 6.7--7.4 tokens per word against roughly 1.3 for English (a); in (b), under
three of the four vocabularies the average Bengali token spans 2.2 - 2.6 bytes,
below the 3-byte Bengali code point (model-weighted mean 2.55; Tekken is the
exception at 4.4), against 4.4 - 4.9 bytes for English.}
    \label{fig:token_stats}
\end{figure*}

\begin{figure*}[t]
    \centering
    \includegraphics[width=0.55\textwidth]{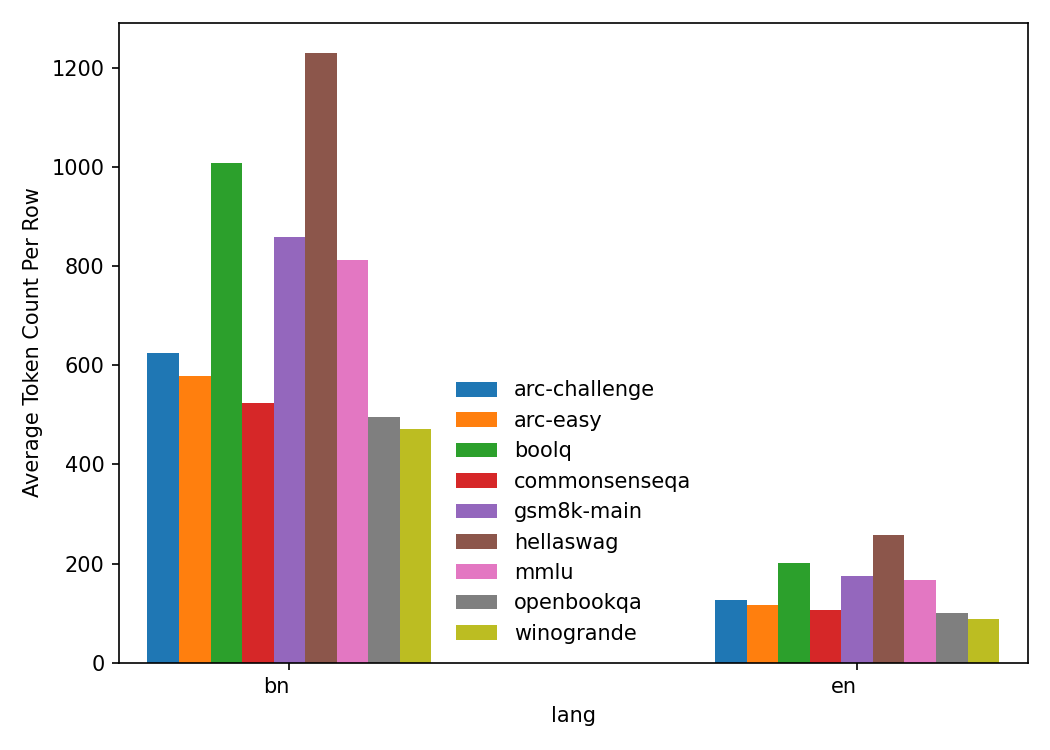}
\caption{Tokens per instance by dataset and language.}
    \label{fig:token_stats_appendix}
\end{figure*}

\begin{figure*}[t]
    \centering
    \begin{subfigure}[t]{0.48\textwidth}
        \centering
        \includegraphics[width=\linewidth]{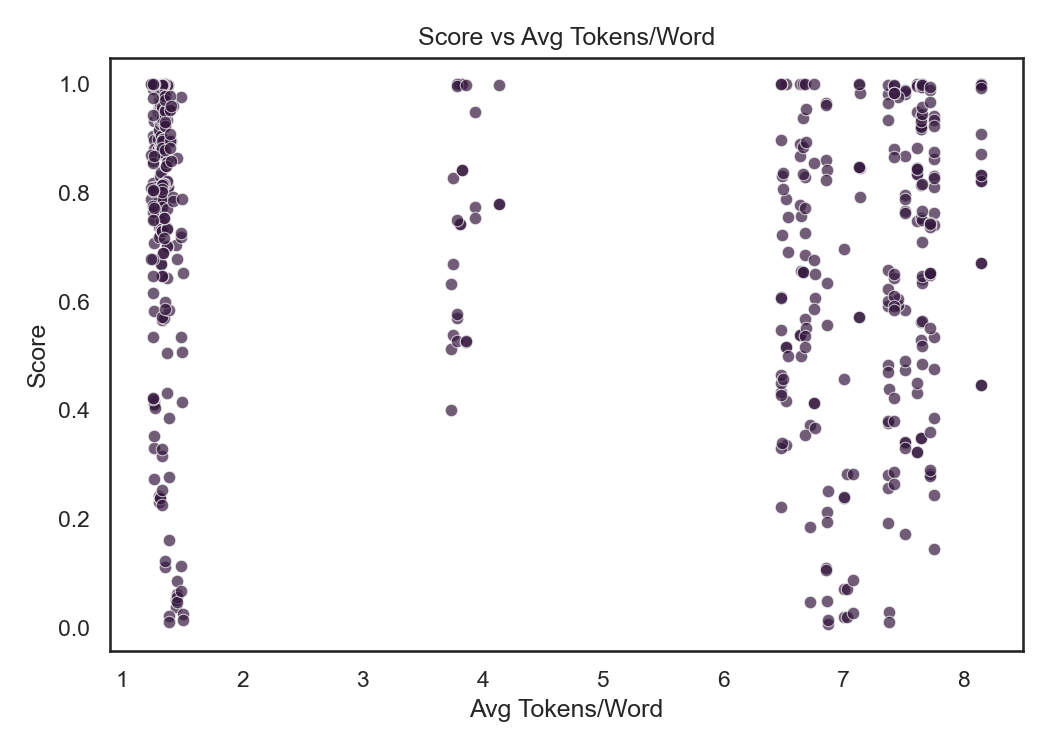}
\caption{Fertility vs.\ LLM-judge score.}
        \label{fig:scatter_fertility}
    \end{subfigure}
    \hfill
    \begin{subfigure}[t]{0.48\textwidth}
        \centering
        \includegraphics[width=\linewidth]{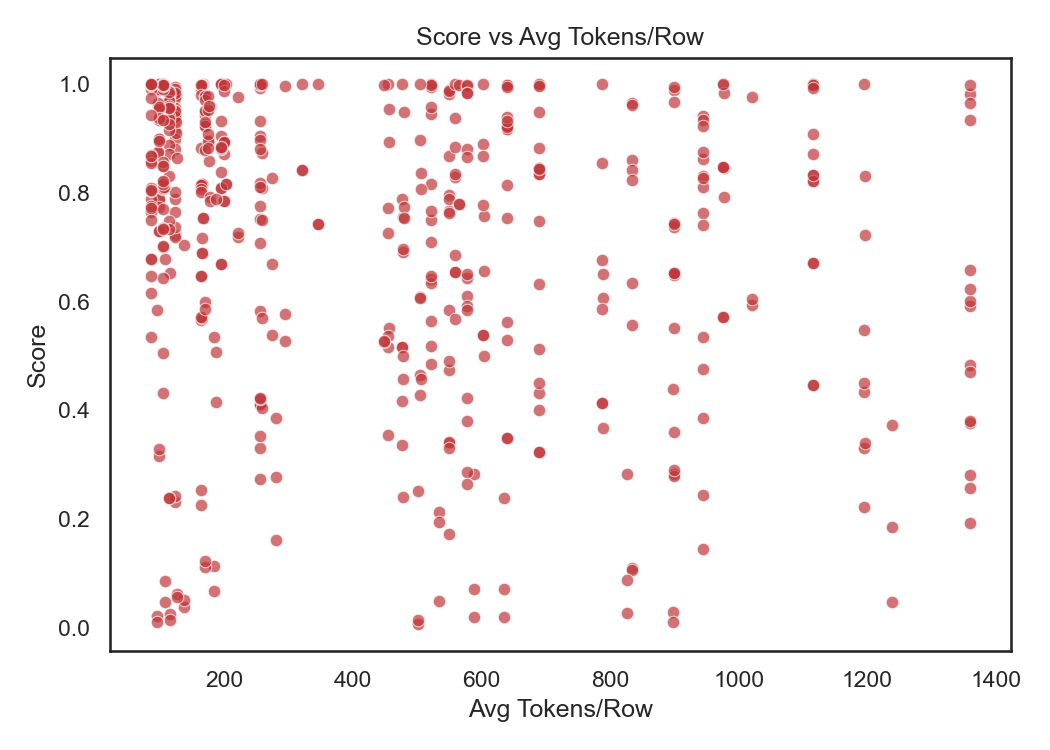}
\caption{Tokens per instance vs.\ LLM-judge score.}
        \label{fig:scatter_atpr}
    \end{subfigure}
\caption{Scatter of tokenizer statistics against LLM-judge scores over
model--dataset cells. Both associations are weak ($r = -0.23$).}
    \label{fig:token_scatter_appendix}
\end{figure*}

The tokens-per-instance columns of Table~\ref{tab:vocab} are pipeline-computed
over the full corpora; its fertility and bytes-per-token columns were measured
by direct tokenization of Bengali text with each vocabulary's public tokenizer
and agree with the pipeline-computed statistics shown here.
Figure~\ref{fig:token_stats_appendix} reports tokens per instance by dataset and
language and Figure~\ref{fig:token_scatter_appendix} shows the scatter of
fertility and tokens-per-instance against LLM-judge scores underlying the
$r=-0.23$ association reported in \S\ref{sec:tokenization}.

\section{Additional Results Figures}
\label{app:results}

\begin{figure}[t]
    \centering
    \includegraphics[width=0.94\linewidth]{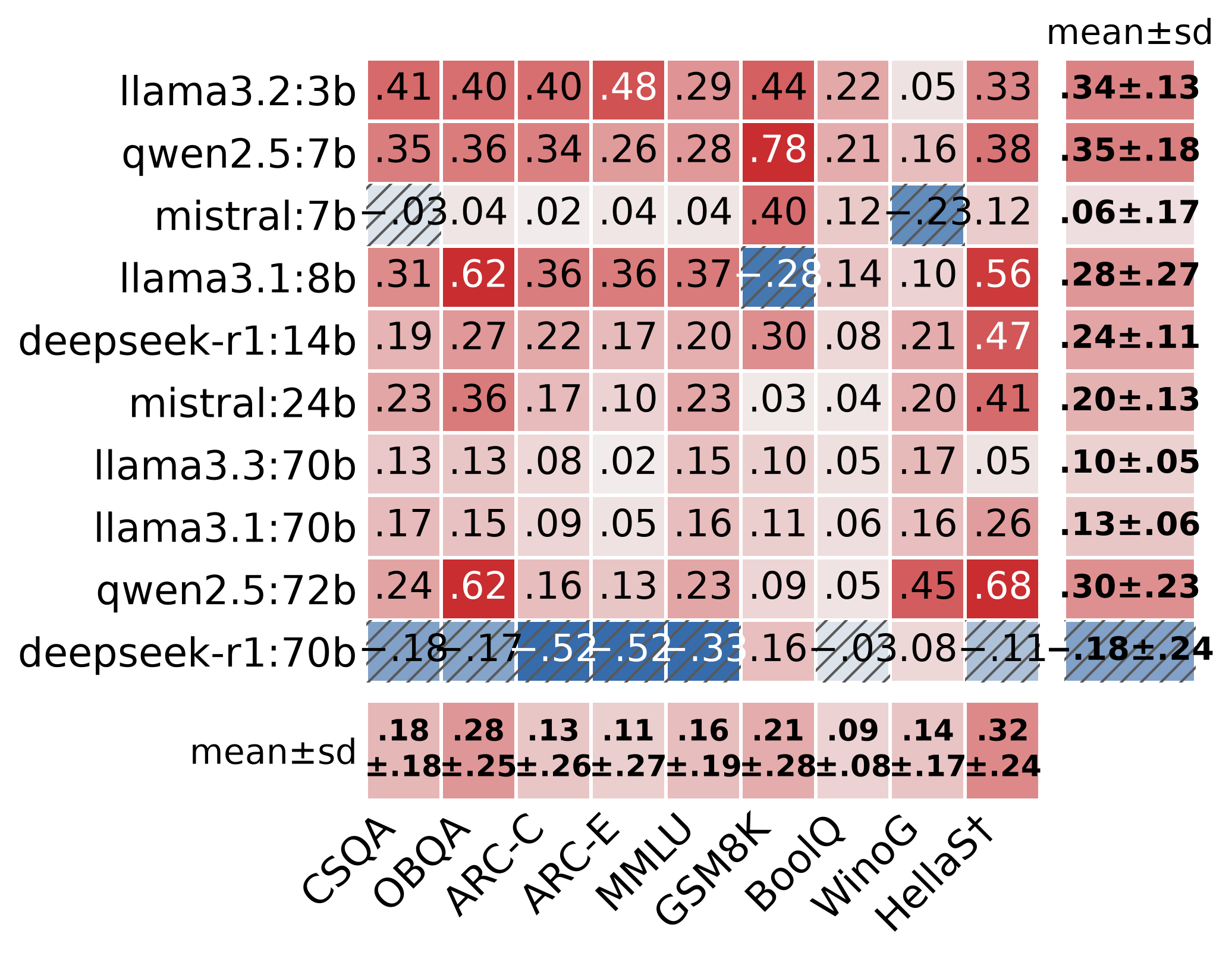}
\caption{Exact-match counterpart of Figure~\ref{fig:score-diff-sorted} with
identical row and column order. The format artifacts of \S\ref{sec:format}
dominate: HellaSwag flips from $-$0.02 (judge) to $+$0.32 mean,
deepseek-r1:70b's row turns negative because format failure suppresses its
English scores and ARC-C/ARC-E swing from consistent judge gaps to standard
deviations of 0.26--0.27.}
    \label{fig:gap_heatmap_em}
\end{figure}

\begin{figure*}[t]
    \centering
    \includegraphics[width=0.8\textwidth]{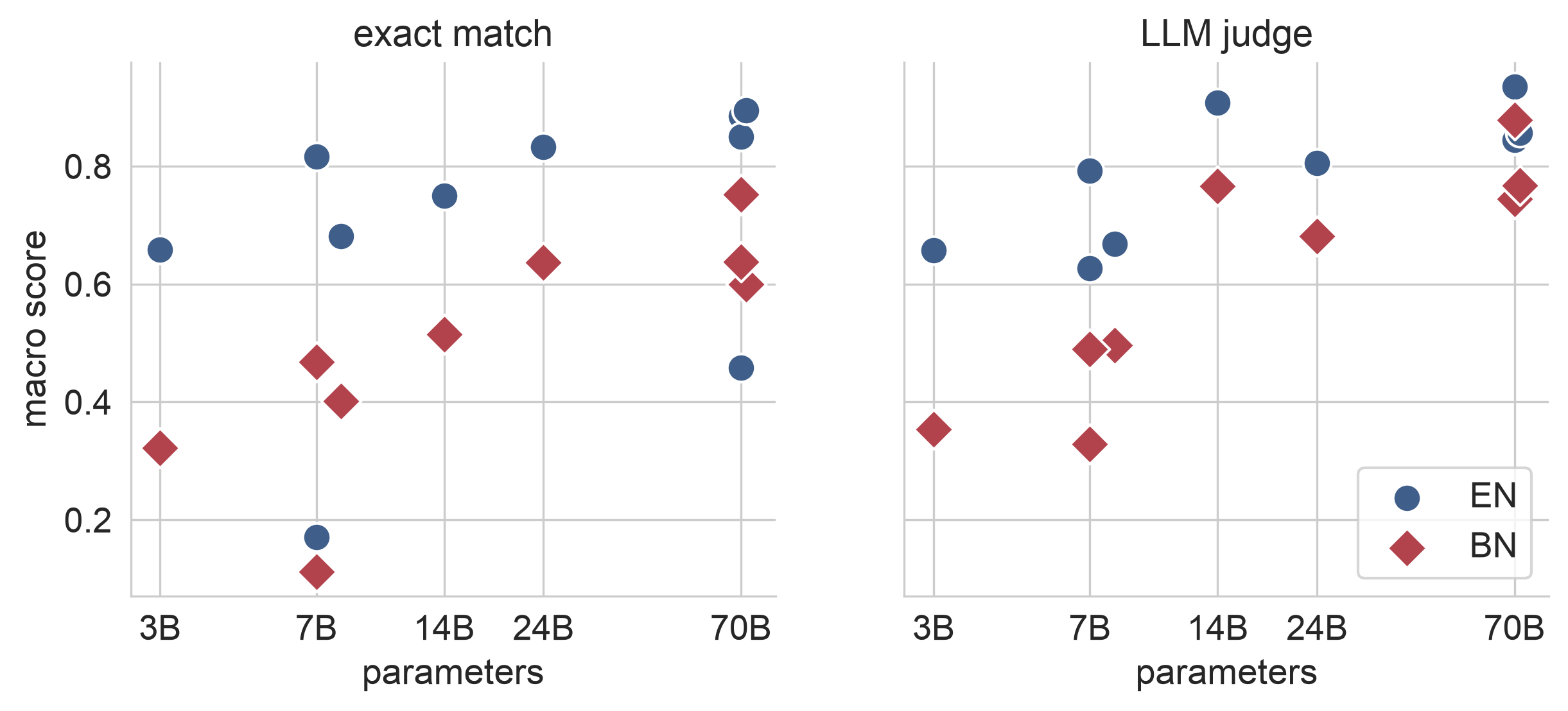}
\caption{Macro scores by model size and language under exact match (left) and
the LLM judge (right). Bengali scores rise faster with scale than English
scores; the exact-match panel additionally shows the format-failure outliers of
\S\ref{sec:format}.}
    \label{fig:model_size_vs_score_by_lang_metric_type}
\end{figure*}

\begin{figure*}[t]
    \centering
    \begin{subfigure}[t]{0.48\textwidth}
        \centering
        \includegraphics[width=\linewidth]{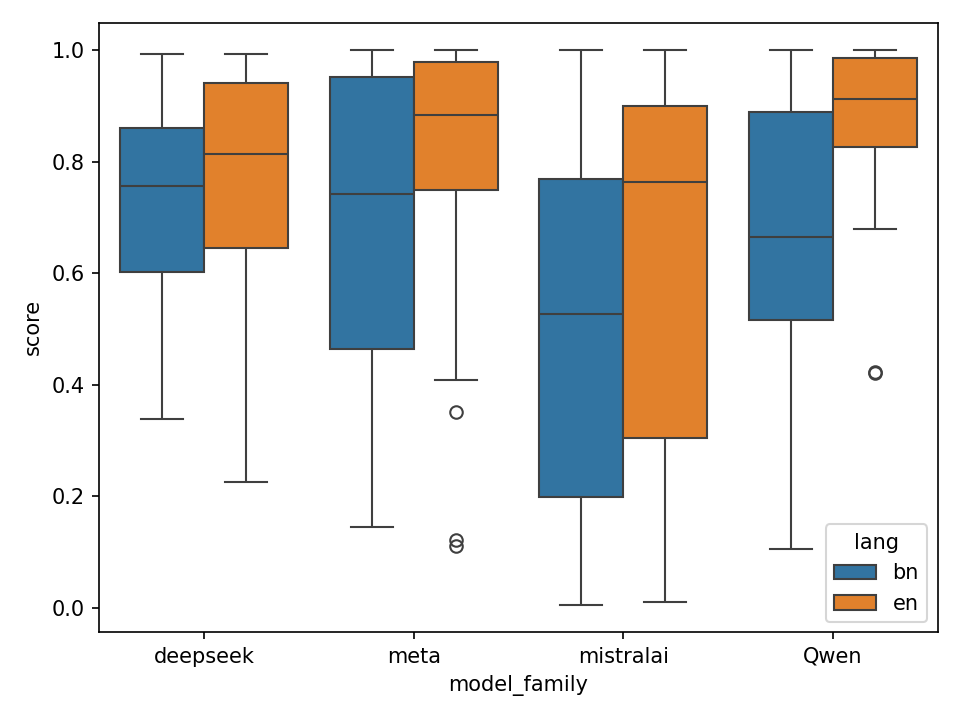}
\caption{LLM-judge score distributions by model family.}
        \label{fig:scores_across_model_family}
    \end{subfigure}
    \hfill
    \begin{subfigure}[t]{0.48\textwidth}
        \centering
        \includegraphics[width=\linewidth]{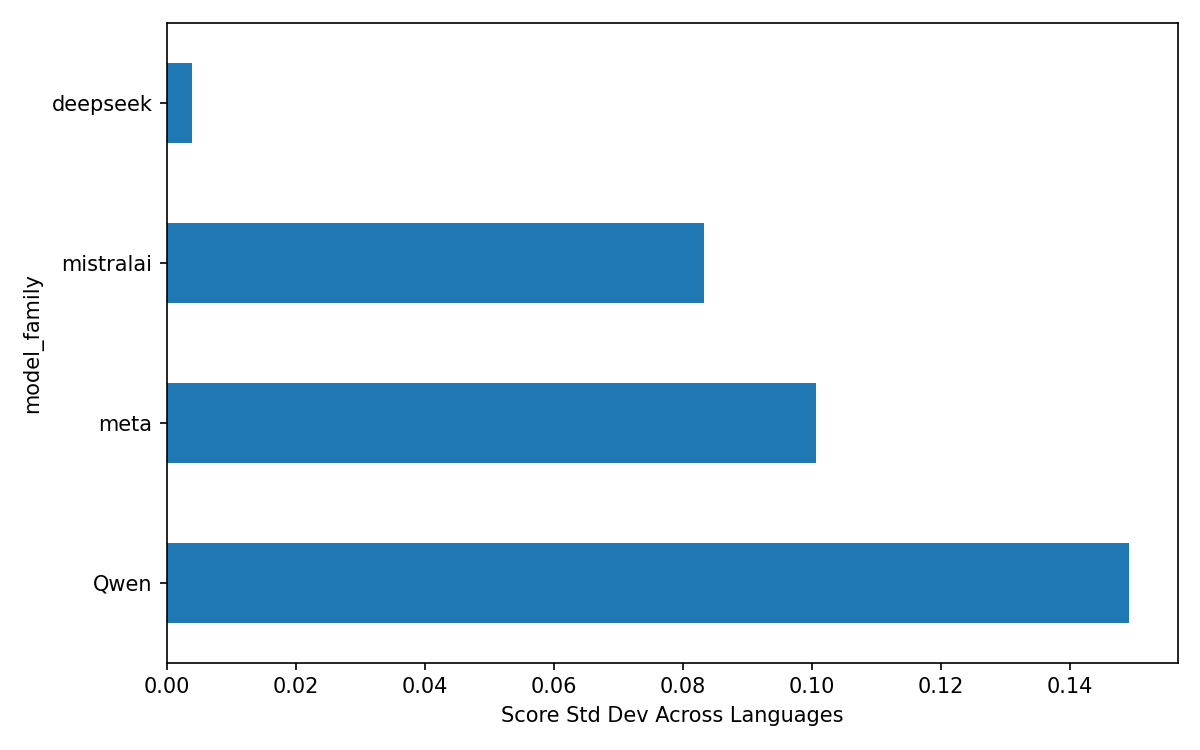}
\caption{Standard deviation of scores across languages by family.}
        \label{fig:score_std_dev_across_lang}
    \end{subfigure}
\caption{Family-level score distributions and cross-language robustness. Lower
deviation in (b) indicates more stable behavior across English and Bengali; the
DeepSeek-R1 family is the most stable, consistent with Table~\ref{tab:macro}.}
    \label{fig:family_appendix}
\end{figure*}

Figure~\ref{fig:family_appendix} shows family-level LLM-judge score
distributions and the standard deviation of scores across languages,
complementing the per-model macro-averages in Table~\ref{tab:macro}.
Figure~\ref{fig:model_size_vs_score_by_lang_metric_type} plots macro scores by
model size and language under both metrics.
Tables~\ref{tab:en_bn_accuracy}--\ref{tab:en_bn_judge} give the per-split scores
underlying that table.

\begin{table*}[t]
\centering
\footnotesize
\setlength{\tabcolsep}{4.0pt}
\begin{tabular}{ll rrrrrrrrr}
\toprule
& \textbf{Model} & OBQA & CSQA & ARC-E & ARC-C & BoolQ & GSM8K & WinoG & HellaS & MMLU \\
\midrule
\multirow{10}{*}{\rotatebox{90}{\textbf{English}}}
& llama3.1:8b     & 0.790 & 0.735 & 0.888 & 0.788 & 0.809 & 0.111 & 0.616 & 0.753 & 0.647 \\
& llama3.1:70b    & 0.938 & 0.816 & 0.969 & 0.934 & 0.882 & 0.923 & 0.805 & 0.882 & 0.814 \\
& llama3.2:3b     & 0.730 & 0.701 & 0.832 & 0.720 & 0.669 & 0.586 & 0.535 & 0.583 & 0.567 \\
& llama3.3:70b    & 0.896 & 0.771 & 0.941 & 0.916 & 0.885 & 0.931 & 0.804 & 0.709 & 0.802 \\
& qwen2.5:7b      & 0.874 & 0.817 & 0.917 & 0.881 & 0.786 & 0.882 & 0.679 & 0.819 & 0.690 \\
& qwen2.5:72b     & 0.960 & 0.849 & 0.969 & 0.943 & 0.893 & 0.909 & 0.809 & 0.905 & 0.817 \\
& mistral:7b      & 0.048 & 0.014 & 0.056 & 0.038 & 0.719 & 0.416 & 0.011 & 0.162 & 0.068 \\
& mistral:24b     & 0.900 & 0.811 & 0.937 & 0.911 & 0.817 & 0.785 & 0.773 & 0.810 & 0.754 \\
& deepseek-r1:14b & 0.774 & 0.645 & 0.733 & 0.723 & 0.872 & 0.859 & 0.766 & 0.811 & 0.571 \\
& deepseek-r1:70b & 0.316 & 0.432 & 0.238 & 0.231 & 0.839 & 0.923 & 0.647 & 0.273 & 0.226 \\
\midrule
\multirow{10}{*}{\rotatebox{90}{\textbf{Bengali}}}
& llama3.1:8b     & 0.172 & 0.423 & 0.529 & 0.433 & 0.671 & 0.387 & 0.519 & 0.193 & 0.282 \\
& llama3.1:70b    & 0.790 & 0.650 & 0.922 & 0.846 & 0.822 & 0.811 & 0.648 & 0.624 & 0.650 \\
& llama3.2:3b     & 0.330 & 0.287 & 0.349 & 0.323 & 0.446 & 0.145 & 0.485 & 0.257 & 0.280 \\
& llama3.3:70b    & 0.764 & 0.643 & 0.918 & 0.835 & 0.833 & 0.827 & 0.635 & 0.659 & 0.652 \\
& qwen2.5:7b      & 0.516 & 0.464 & 0.654 & 0.538 & 0.572 & 0.106 & 0.516 & 0.435 & 0.414 \\
& qwen2.5:72b     & 0.336 & 0.609 & 0.835 & 0.779 & 0.848 & 0.824 & 0.355 & 0.222 & 0.587 \\
& mistral:7b      & 0.006 & 0.048 & 0.019 & 0.019 & 0.594 & 0.011 & 0.240 & 0.046 & 0.026 \\
& mistral:24b     & 0.538 & 0.577 & 0.842 & 0.743 & 0.780 & 0.754 & 0.570 & 0.401 & 0.527 \\
& deepseek-r1:14b & 0.500 & 0.457 & 0.568 & 0.500 & 0.792 & 0.557 & 0.552 & 0.339 & 0.367 \\
& deepseek-r1:70b & 0.490 & 0.611 & 0.755 & 0.749 & 0.872 & 0.764 & 0.565 & 0.381 & 0.552 \\
\bottomrule
\end{tabular}
\caption{Exact-match accuracy per evaluation split, English (top block) and
Bengali (bottom block).}
\label{tab:en_bn_accuracy}
\end{table*}

\begin{table*}[t]
\centering
\footnotesize
\setlength{\tabcolsep}{4.0pt}
\begin{tabular}{ll rrrrrrrrr}
\toprule
& \textbf{Model} & OBQA & CSQA & ARC-E & ARC-C & BoolQ & GSM8K & WinoG & HellaS & MMLU \\
\midrule
\multirow{10}{*}{\rotatebox{90}{\textbf{English}}}
& llama3.1:8b     & 0.000 & 0.001 & 0.040 & 0.015 & 0.000 & 0.121 & 0.000 & 0.000 & 0.000 \\
& llama3.1:70b    & 0.010 & 0.008 & 0.013 & 0.005 & 0.000 & 0.022 & 0.002 & 0.000 & 0.001 \\
& llama3.2:3b     & 0.000 & 0.003 & 0.041 & 0.019 & 0.000 & 0.073 & 0.000 & 0.000 & 0.018 \\
& llama3.3:70b    & 0.042 & 0.065 & 0.043 & 0.026 & 0.000 & 0.022 & 0.000 & 0.223 & 0.009 \\
& qwen2.5:7b      & 0.008 & 0.000 & 0.041 & 0.018 & 0.000 & 0.020 & 0.000 & 0.000 & 0.001 \\
& qwen2.5:72b     & 0.008 & 0.000 & 0.019 & 0.009 & 0.001 & 0.045 & 0.000 & 0.000 & 0.000 \\
& mistral:7b      & 0.914 & 0.975 & 0.938 & 0.950 & 0.023 & 0.211 & 0.979 & 0.722 & 0.886 \\
& mistral:24b     & 0.000 & 0.002 & 0.039 & 0.020 & 0.000 & 0.039 & 0.000 & 0.125 & 0.000 \\
& deepseek-r1:14b & 0.190 & 0.192 & 0.250 & 0.235 & 0.011 & 0.039 & 0.056 & 0.007 & 0.282 \\
& deepseek-r1:70b & 0.672 & 0.494 & 0.761 & 0.757 & 0.067 & 0.028 & 0.249 & 0.669 & 0.747 \\
\midrule
\multirow{10}{*}{\rotatebox{90}{\textbf{Bengali}}}
& llama3.1:8b     & 0.658 & 0.016 & 0.061 & 0.051 & 0.000 & 0.258 & 0.005 & 0.407 & 0.257 \\
& llama3.1:70b    & 0.012 & 0.003 & 0.002 & 0.001 & 0.002 & 0.065 & 0.000 & 0.001 & 0.005 \\
& llama3.2:3b     & 0.018 & 0.002 & 0.001 & 0.000 & 0.000 & 0.465 & 0.041 & 0.018 & 0.032 \\
& llama3.3:70b    & 0.010 & 0.016 & 0.004 & 0.003 & 0.000 & 0.058 & 0.001 & 0.035 & 0.009 \\
& qwen2.5:7b      & 0.000 & 0.000 & 0.000 & 0.000 & 0.000 & 0.039 & 0.000 & 0.000 & 0.000 \\
& qwen2.5:72b     & 0.582 & 0.102 & 0.116 & 0.110 & 0.000 & 0.033 & 0.463 & 0.670 & 0.144 \\
& mistral:7b      & 0.986 & 0.788 & 0.930 & 0.929 & 0.024 & 0.561 & 0.542 & 0.815 & 0.913 \\
& mistral:24b     & 0.172 & 0.003 & 0.000 & 0.000 & 0.000 & 0.051 & 0.000 & 0.367 & 0.000 \\
& deepseek-r1:14b & 0.244 & 0.162 & 0.314 & 0.344 & 0.015 & 0.158 & 0.046 & 0.277 & 0.348 \\
& deepseek-r1:70b & 0.416 & 0.118 & 0.186 & 0.156 & 0.007 & 0.077 & 0.184 & 0.399 & 0.256 \\
\bottomrule
\end{tabular}
\caption{Response Error Rate (RER, lower is better) per evaluation split,
English (top block) and Bengali (bottom block). Note the strong model and
language asymmetries: mistral:7b fails format in both languages, deepseek-r1:70b
mainly in English, llama3.1:8b and qwen2.5:72b mainly in Bengali.}
\label{tab:en_bn_rer}
\end{table*}

\begin{table*}[t]
\centering
\footnotesize
\setlength{\tabcolsep}{4.0pt}
\begin{tabular}{ll rrrrrrrrr}
\toprule
& \textbf{Model} & OBQA & CSQA & ARC-E & ARC-C & BoolQ & GSM8K & WinoG & HellaS & MMLU \\
\midrule
\multirow{10}{*}{\rotatebox{90}{\textbf{English}}}
& llama3.1:8b     & 0.790 & 0.735 & 0.928 & 0.801 & 0.809 & 0.122 & 0.777 & 0.409 & 0.648 \\
& llama3.1:70b    & 0.940 & 0.820 & 0.982 & 0.939 & 0.882 & 0.945 & 0.856 & 0.412 & 0.814 \\
& llama3.2:3b     & 0.730 & 0.703 & 0.872 & 0.738 & 0.669 & 0.600 & 0.680 & 0.352 & 0.572 \\
& llama3.3:70b    & 0.934 & 0.822 & 0.983 & 0.941 & 0.885 & 0.953 & 0.860 & 0.417 & 0.807 \\
& qwen2.5:7b      & 0.876 & 0.817 & 0.957 & 0.899 & 0.786 & 0.896 & 0.789 & 0.423 & 0.690 \\
& qwen2.5:72b     & 0.966 & 0.849 & 0.988 & 0.951 & 0.893 & 0.952 & 0.870 & 0.421 & 0.817 \\
& mistral:7b      & 0.678 & 0.654 & 0.865 & 0.704 & 0.727 & 0.508 & 0.585 & 0.387 & 0.535 \\
& mistral:24b     & 0.900 & 0.811 & 0.977 & 0.930 & 0.817 & 0.792 & 0.869 & 0.404 & 0.754 \\
& deepseek-r1:14b & 0.958 & 0.810 & 0.980 & 0.948 & 0.893 & 0.895 & 0.976 & 0.898 & 0.816 \\
& deepseek-r1:70b & 0.954 & 0.857 & 0.987 & 0.957 & 0.906 & 0.951 & 0.991 & 0.933 & 0.883 \\
\midrule
\multirow{10}{*}{\rotatebox{90}{\textbf{Bengali}}}
& llama3.1:8b     & 0.474 & 0.381 & 0.562 & 0.451 & 0.671 & 0.477 & 0.710 & 0.377 & 0.360 \\
& llama3.1:70b    & 0.796 & 0.593 & 0.923 & 0.846 & 0.823 & 0.862 & 0.768 & 0.470 & 0.652 \\
& llama3.2:3b     & 0.340 & 0.265 & 0.350 & 0.323 & 0.446 & 0.243 & 0.642 & 0.282 & 0.290 \\
& llama3.3:70b    & 0.768 & 0.585 & 0.918 & 0.835 & 0.833 & 0.876 & 0.750 & 0.484 & 0.653 \\
& qwen2.5:7b      & 0.516 & 0.429 & 0.654 & 0.538 & 0.572 & 0.110 & 0.726 & 0.451 & 0.414 \\
& qwen2.5:72b     & 0.790 & 0.606 & 0.938 & 0.868 & 0.848 & 0.861 & 0.772 & 0.547 & 0.677 \\
& mistral:7b      & 0.252 & 0.195 & 0.283 & 0.239 & 0.605 & 0.030 & 0.697 & 0.372 & 0.282 \\
& mistral:24b     & 0.670 & 0.528 & 0.842 & 0.743 & 0.780 & 0.775 & 0.751 & 0.514 & 0.527 \\
& deepseek-r1:14b & 0.692 & 0.808 & 0.830 & 0.758 & 0.847 & 0.634 & 0.895 & 0.832 & 0.606 \\
& deepseek-r1:70b & 0.868 & 0.866 & 0.933 & 0.883 & 0.909 & 0.831 & 0.946 & 0.934 & 0.737 \\
\bottomrule
\end{tabular}
\caption{LLM-judge score per evaluation split, English (top block) and Bengali
(bottom block).}
\label{tab:en_bn_judge}
\end{table*}

\section{Reproducibility Details}
\label{app:repro}

\begin{table}[t]
  \centering
  \resizebox{\columnwidth}{!}{%
    \begin{tabular}{@{} l c c c @{}}
      \toprule
      Model & Size & Multiling. & Bengali pretraining \\
      \midrule
      LLaMA 3.1 \citep{grattafiori2024llama3herdmodels}  & 8B   & Limited & \xmark \\
      LLaMA 3.1 \citep{grattafiori2024llama3herdmodels}  & 70B  & Limited & \xmark \\
      LLaMA 3.2 \citep{grattafiori2024llama3herdmodels}  & 3B   & Limited & \xmark \\
      LLaMA 3.3 \citep{grattafiori2024llama3herdmodels}  & 70B  & Limited & \xmark \\
      Qwen 2.5 \citep{qwen2025qwen25technicalreport}     & 7B   & Yes     & \cmark \\
      Qwen 2.5 \citep{qwen2025qwen25technicalreport}     & 72B  & Yes     & \cmark \\
      Mistral \citep{jiang2023mistral7b}                 & 7B   & No      & \xmark \\
      Mistral Small \citep{mistral-small-24b-instruct-2501} & 24B & Yes   & \xmark \\
      DeepSeek-R1 Distill--Qwen \citep{guo2025deepseek}  & 14B  & Yes     & \cmark \\
      DeepSeek-R1 Distill--Llama \citep{guo2025deepseek} & 70B  & Limited & \xmark \\
      \bottomrule
    \end{tabular}
  }
\caption{Evaluated models (chat/instruct variants). Bengali coverage follows
each model's documentation. The DeepSeek-R1 models are reasoning distillations
onto Qwen~2.5 (14B) and Llama~3.3 (70B) bases and inherit those bases' language
coverage.}
  \label{tab:models}
\end{table}

\begin{figure*}[t]
    \centering
    \includegraphics[width=0.55\textwidth]{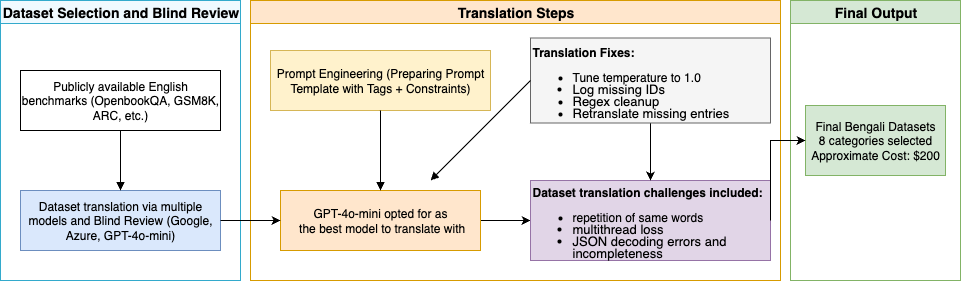}
\caption{Overview of the construction and evaluation pipeline: benchmark
selection, a blind three-way translator comparison, GPT-4o-mini translation with
JSON-constrained outputs, automated validation with retranslation and paired
English--Bengali evaluation of 10 open LLMs under three metrics.}
    \label{fig:methodology_overview}
\end{figure*}

Figure~\ref{fig:methodology_overview} gives a schematic overview of the
construction and evaluation pipeline.

\begin{table}[ht]
  \centering
  \small
  \begin{tabularx}{\columnwidth}{>{\bfseries}l X}
    \toprule
    Role   & Content \\
    \midrule
    System & You are a professional translator tasked with accurately translating text from English to Bengali. Your primary goal is to provide precise and culturally appropriate translations, regardless of the content's nature. \\
    \midrule
    User   & Translate the following English text into Bengali and ensure the output is valid JSON with all strings enclosed in double quotes: \texttt{<english\_text>} \texttt{\{\{"input": \{input\}, "target": \{target\}\}\}} \texttt{</english\_text>} Guidelines: 1.~Translate accurately, maintaining meaning, tone and context. 2.~Handle idiomatic expressions appropriately. 3.~Preserve specialized terminology and proper nouns. 4.~Translate sensitive content accurately without censorship. 5.~Do not translate JSON keys, only values. 6.~Ensure valid JSON output with double-quoted strings. Output within \texttt{<bangla\_translation>} tags; notes in \texttt{<translator\_notes>} tags. \\
    \bottomrule
  \end{tabularx}
\caption{Prompt template for English-to-Bengali translation.}
  \label{tab:translation-prompt}
\end{table}
All models were evaluated zero-shot using their chat/instruct variants. Prompts
instruct the model to output the answer as \texttt{Correct Answer: [option]}
(with a Bengali-language equivalent instruction for Bengali inputs) and
responses were scored as described in \S\ref{sec:metrics}. Exact-match
normalization lowercases responses and maps label variants (e.g., ``Yes'' to
``yes'') before prefix comparison. The LLM judge is a few-shot GPT-based
classifier that receives the question, ground truth and model response and
returns a Correct/Incorrect verdict. Evaluation uses each benchmark's released
evaluation split as listed in Table~\ref{tab:inventory} (the validation split
where test labels are not public); exact per-cell counts ship with the released
datasets. Tokenization statistics were computed over the concatenation of system
prompt and item for every instance in each dataset, per model tokenizer. The
released repository contains the translation pipeline, all prompt templates in
both languages, inference scripts, judge prompts and the scripts that produce
every table and figure in this paper. Datasets are released with train/dev/test
splits mirroring their English sources.

\vspace{0.5em}
\noindent\textbf{Artifact availability:} The translated datasets, translation pipeline and evaluation code are publicly released: code on \href{https://github.com/BengaliAI/bn-llm-benchmark}{GitHub} and datasets in our \href{https://huggingface.co/collections/bengaliAI/bengali-llm-benchmark-datasets-683bd5999bb4c70bc9e83310}{Hugging Face collection}.

\end{document}